\documentclass[review]{elsarticle}

\usepackage{hyperref}

\journal{Journal of Information Processing and Management}

\biboptions{authoryear}

\usepackage{CJKutf8}
\newcommand*{\Ja}[1]{\begin{CJK}{UTF8}{ipxm}#1\end{CJK}}

\usepackage{array}
\usepackage{amssymb}

\date{}

\begin{document}

\begin{frontmatter}

  \title{News-based Business Sentiment and its \\Properties as an Economic Index}


\author[konan]{Kazuhiro Seki\corref{mycorrespondingauthor}}
\cortext[mycorrespondingauthor]{Corresponding author: Kazuhiro Seki}
\ead[url]{http://www.konan-u.ac.jp/hp/seki/}
\ead{seki@konan-u.ac.jp}

\author[sangyo]{Yusuke Ikuta}

\author[kobe,apir]{Yoichi Matsubayashi}

\address[konan]{{\normalsize Konan University}\\8-9-1 Okamoto, Higashinada, Kobe, Hyogo 658-8501, Japan}

\address[sangyo]{{\normalsize Osaka Sangyo University}\\3-1-1 Nakagaito, Daito, Osaka 574-8530, Japan}

\address[kobe]{{\normalsize Kobe University}\\1-1 Rokkodaicho, Nada, Kobe, Hyogo 657-8501, Japan}
\address[apir]{{\normalsize Asia Pacific Institute of Research}\\3-1 Ofuka, Kita, Osaka 530-0011, Japan}

\begin{abstract}
This paper presents an approach to measuring business sentiment based on textual data. Business sentiment has been measured by traditional surveys, which are costly and time-consuming to conduct.  To address the issues, we take advantage of daily newspaper articles and adopt a self-attention-based  model to define a business sentiment index, named S-APIR, where outlier detection models are investigated to properly handle various genres of news articles.  Moreover, we propose a simple approach to temporally analyzing how much any given event contributed to the predicted business sentiment index. To demonstrate the validity of the proposed approach, an extensive analysis is carried out on 12 years' worth of newspaper articles.  The analysis shows that the S-APIR index is strongly and positively correlated with established survey-based index (up to correlation coefficient $r=0.937$) and that the outlier detection is effective especially for a general newspaper.  Also, S-APIR is compared with a variety of economic indices, revealing the properties of S-APIR that it reflects the trend of the macroeconomy as well as the economic outlook and sentiment of economic agents.  Moreover, to illustrate how S-APIR could benefit economists and policymakers, several events are analyzed with respect to their impacts on business sentiment over time.

\end{abstract}


\end{frontmatter}


\section{Introduction}
\label{sec:intro}

In Japan, there exist business sentiment indices, such as Economy Watchers Survey\footnote{\url{https://www5.cao.go.jp/keizai3/watcher-e/index-e.html}} and Short‐term Economic Survey of Principal Enterprise\footnote{\url{https://www.boj.or.jp/en/statistics/tk/long_syu/index.htm/}} conducted by the Government and the Bank of Japan, respectively.  These diffusion indices (DI) play a crucial role in decision-making for governmental/monetary policies, industrial production planning, institutional/private investment, and so on. However, these DIs rely on traditional surveys, which are costly and time-consuming to conduct. 

For example, Economy Watchers Survey is carried out in 12 regions of Japan, where 2,050 preselected respondents who can observe the regional business/economic conditions (e.g., store owners and taxi drivers) fill out a questionnaire and then an investigative organization in each region aggregates the surveys and calculates a DI.  As the survey and subsequent processes take time, the DI is published only monthly.

On the other hand, so-called alternative data, including merchandise sales, news, micro-blogs, query logs, credit card transactions, GPS location information, and satellite images, are constantly generated and accumulated.  The availability of such data has accelerated the development of data-driven artificial intelligence (AI) models and techniques represented by deep learning.  In econometrics, there is a growing interest in future/current forecasts of economic and financial indices by using such alternative, large-scale data instead of traditional surveys~\citep{chen19:_off_races,haughwout19:_macro}. For example, point of sales (POS) data have been used for estimating consumer price index (CPI)~\citep{watanabe14:_estim_daily_inflat_using_scann_data}; financial and economic reports for business sentiment~\citep{yamamoto16eng}; newspaper for stock prices~\citep{li20:_incor,picasso19:_techn,yoshihara14:_predic_stock_market_trend_recur,yoshihara16:_lever}, socio-economic indicators~\citep{chakraborty16:_predic_socio_econom_indic_using_news_event},  consumer sentiment~\citep{shapiro20:_measur}; and social media for stock prices~\citep{bollen11:_twitt,derakhshan19:_sentim,levenberg14:_predic_econom_indic_web_text}.

This work focuses on textual data and uses daily newspaper articles to develop a new business sentiment index, named the S-APIR index.  In addition, using the computed index, we propose an approach to temporally analyzing the influence of an arbitrary event on business sentiment.

The remainder of the paper is structured as follows: Section~\ref{sec:related_work} introduces the related work on sentiment analysis in general and its applications to market sentiment and business sentiment prediction.  Section~\ref{sec:research-objectives} states the research objectives pursued in the present work.  Section~\ref{sec:s-apir} details our proposed approach to forecasting business sentiment index and describes how to temporally analyze the contribution of a given event to business sentiment index based on predicted business sentiment scores.  Section~\ref{sec:evlaluation} conducts evaluative experiments using over 12 years' worth of newspaper articles and discusses the properties of S-APIR, in addition to word-level temporal analysis. Section~\ref{sec:implications} discusses the implications and findings of this work. Section~\ref{sec:conclusions} concludes with a brief summary and possible future directions.

\section{Related Work}
\label{sec:related_work}

In the economic and financial domains, there are abundant textual data, such as newspaper articles and financial reports as well as many numerical data. These texts are intended to be read by humans, who also consider other sources of information and make decisions on investment, financial policies, and so on. However, it is difficult even for experts to read and grasp all the available information in a limited time. Therefore, there has been much research on computing economical/financial indices from textual data, which is closely related to \textit{sentiment analysis}.  In the following, we briefly introduce the related work in general sentiment analysis, and then summarize its applications to market sentiment prediction. Finally, we describe the related work in business sentiment prediction, which is the main focus of the present work.
 
\subsection{Sentiment Analysis}

Sentiment analysis is a sub-field of natural language processing (NLP) and aims to predict the sentiment orientation (i.e., positive or negative) or sentiment score for a given text~\citep{yadollahi17:_curren_state_text_sentim_analy}.  Sentiment analysis is often applied to user-generated content such as tweets~\citep{giachanou16:_like_it_not,zimbra18:_state_art_twitt_sentim_analy} and product reviews~\citep{fang15:_sentim} to understand the opinions of users about an object (e.g., product, service, person, and company), which can be valuable information for, for example, improving products for manufacturers and making decisions for customers.  Note that it is also possible to consider multiple categories of sentiment, such as happy, sad, angry, and embarrassed, instead of the dichotomous, positive/negative sentiment.

Sentiment analysis approaches can be roughly categorized into lexicon-based~\citep{khoo18:_lexic}, rule-based~\citep{vilares17:_univer}, and machine learning-based~\citep{kouadri20:_qualit_sentim_analy_tools}. Lexicon-based approaches use lists of words, each belonging to a sentiment category, and look for the occurrences of the words in the list.  Rule-based approaches are similar to lexicon-based but add another layer of inference based on linguistic rules.  Machine learning-based approaches employ supervised or semi-supervised learning models to classify a given text into predefined sentiment categories.  As supervised learning models, deep learning models have been popularly used in recent years, including memory neural network (MNN)~\citep{tay17:_dyadic_memor_networ_aspec_based_sentim_analy}, recurrent neural network (RNN) with long short-term memory (LSTM) units~\citep{song19:_atten_korean,xu19:_sentim_analy_commen_texts_based_bilst}, combination of LSTM and convolutional neural network (CNN)~\citep{behera21:_co_lstm,rehman19:_hybrid_cnn_lstm_model_improv}, and attention-based language representation models~\citep{farha21:_arabic,pota21:_multil_bert,smetanin21:_deep_russian}.

\subsection{Market Sentiment Prediction}

Market sentiment prediction is an application of sentiment analysis techniques to the stock market domain.  Seminal work in market sentiment prediction was conducted by \cite{bollen11:_twitt}.  They collected microblog posts (i.e., tweets) from Twitter and applied a lexicon-based sentiment analysis using six categories: calm, alert, sure, vital, kind, and happy.  Their analysis showed that the calm score and Dow Jones Industrial Average have a causal relation.  Since Bollen et al's work, there has been much work utilizing social media for forecasting stock market indices~\citep{arias14:_forec_twitt_data,oliveira17}, predicting stock price (or movement)~\citep{derakhshan19:_sentim,li17:_discov,nguyen17:_distin_anton_synon_patter_neural_networ,tu18:_inves}, investigating the effects of users' emotions on the stock market during a market crash~\citep{ge20:_beyon}, and analyzing the impact of bullish-bearish tendencies estimated from the online financial community forum on market volatility and market returns~\citep{qian20}.  \cite{ren21:_how} analyzed the interaction between social media sentiment and mass media sentiment.

For predicting stock prices or their movements, the sentiment of financial news texts has been also utilized~\citep{li14:_news}. \cite{zhang18:_improv} jointly used news texts and social media considering their interaction via matrix factorization.  \cite{li20:_incor} and \cite{picasso19:_techn} independently proposed to combine the news sentiment and technical analysis for improving prediction.

\subsection{Business Sentiment Prediction}

Similar to market sentiment prediction, sentiment analysis techniques have been applied to business sentiment prediction~\citep{shapiro20:_measur}.  For Japanese data, Economy Watchers Survey mentioned in Section~\ref{sec:intro} is often used as training data to learn a supervised prediction model.  Since our study also relies on the Economy Watchers Survey, we first describe the data.

Economy Watchers Survey publishes not only the business sentiment index (hereafter called EWDI) but also individual survey responses on which EWDI is based. The survey responses contain a pair of economic condition on a five-point scale from ``$\circledcirc$'' (good) to ``$\times$'' (bad) and a statement of the reason(s) why the respondent chose a particular economic condition.  Some examples of responses translated to English are shown in Table~\ref{tab:keiki_watcher}. Based on the responses, EWDI is computed by first computing the composition ratios of the five economic conditions ($\times$, $\vartriangle$, \scalebox{0.8}{$\square$}, \raisebox{0.3mm}{\scriptsize$\bigcirc$}, $\circledcirc$) and then taking their weighted sum. Resulting EWDI ranges from 0 to 100 with 0 being negative and 100 being positive.

\begin{table}[tb]
  \small
  \caption{Example responses from Economy Watchers Survey
    (translated).}
  \label{tab:keiki_watcher}
  \begin{tabular}{>{\centering\arraybackslash} m{12mm}
    >{\centering\arraybackslash} m{26mm} >{\centering\arraybackslash}
    m{16mm} m{50mm}}
    \hline
    \multicolumn{1}{c}{Region} &
                                 \multicolumn{1}{c}{Occupation}&Economic condition
    & \multicolumn{1}{c}{Statement of reason(s)}\\
    \hline
    Hokkaido & Taxi driver & \multicolumn{1}{c}{$\times$} &  Although sales are declining,
                                                            seasonal factors and the
                                                            downturn in the economy are
                                                            also affecting.\\
    North Kanto & Transportation machinery and equipment manufacturing
                                                               & \multicolumn{1}{c}{$\circledcirc$} & Automobile exports to the United States are increasing.\\
    \hline
  \end{tabular}
\end{table}

\cite{yamamoto16eng} used approximately 200,000 pairs of an economic condition and statement of the reason(s) to learn a bi-directional RNN with LSTM~\citep{hochreiter97:_long_short_term_memor} in order to predict the business sentiment of a given text.  Then, monthly economic reports were fed to the learned model to compute a business sentiment index. They reported that the computed index was positively correlated with both EWDI and the Short‐term Economic Survey of Principal Enterprise in Japan.

\cite{aiba18eng} used a similar model to compute a business sentiment index from micro-blogs (tweets), \cite{seki20:_s_apir} from newspaper articles, and \cite{kondo19eng} from bank's internal documents written from interviews with their client corporations.  \cite{goshima19eng} adopted a CNN and used Reuters news articles to compute a business sentiment index.  \cite{yono17eng} additionally used Latent Dirichlet Allocation~\citep{Blei:2003:LDA:944919.944937} to analyze latent topics and discussed their contributions to the estimated sentiment index.  Following up our work~\citep{seki20:_s_apir}, we applied the Bidirectional Encoder Representation from Transformers (BERT) to measure business sentiment~\citep{seki21:_nowcas_busin_sentim_econom_news_artic}, which was found to be strongly positively correlated with survey-based EWDI, outperforming the related work.  However, the properties of the predicted business sentiment or its usefulness have not been fully discussed in our preceding paper, which motivated the present work.  This paper aims at shedding light on the properties of S-APIR as an economic index and to demonstrate its practical values as discussed in the next section.

\section{Research Objectives}
\label{sec:research-objectives}

This paper is an extension of our previous work~\citep{seki21:_nowcas_busin_sentim_econom_news_artic} with more focus on studying the properties of our business sentiment index, S-APIR, and its application to the temporal analysis of arbitrary events.  Specifically, the research objectives (RO) of this work are the followings:

\begin{itemize}
\item RO1: To measure business sentiment based on daily newspaper articles and empirically validate the proposed approach.
\item RO2: To analyze the properties of the S-APIR index in comparison with various representative economic indicators and discuss its implications.
\item RO3: To quantify the effects of several notable events on business sentiment and illustrate how it could benefit economists and policymakers.
\end{itemize}

To accomplish these goals, we devise a framework to measure business sentiment from newspaper articles based on supervised machine learning models and conduct extensive experiments.  Also, to reveal the properties of the S-APIR index, we systematically compare it against a number of macroeconomic indicators and semi-macro indicators including sentiment indices and actual activity indices.  Lastly, we present and discuss the result of temporal analysis of significant events that occurred between 2008 to 2020.

\section{Turning News Texts into Business Sentiment Index}
\label{sec:s-apir}

\subsection{Overview}
\label{sec:overview}

\begin{figure}[tb]
  \centering
  \includegraphics[width=.75\linewidth]{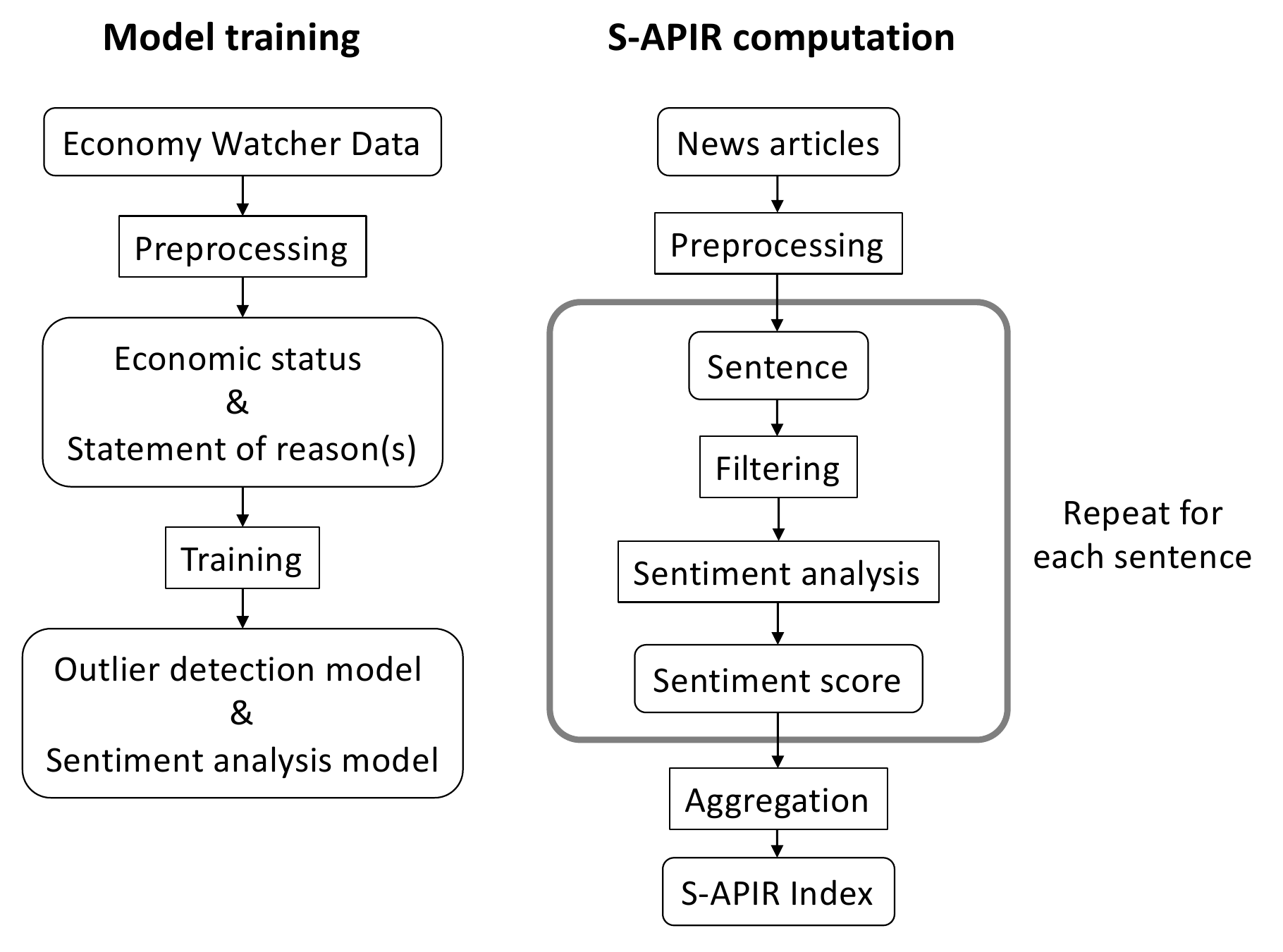}
  \caption{An overview of the approach to predicting a business
    sentiment index.}
  \label{fig:overview}
\end{figure}

This section describes our approach~\citep{seki21:_nowcas_busin_sentim_econom_news_artic} to predicting business sentiment based on news articles.  The overview of the approach is illustrated in Figure~\ref{fig:overview}, where the left-hand side of the figure corresponds to model training and the right-hand side shows the flow of the processes to compute the business sentiment index.  The following sections detail the major components of the approach.

\subsection{Model Training}
\label{sec:model-training}

\subsubsection{Sentiment Analysis Model}
\label{sec:finetune}

Following the related work~\citep{yamamoto16eng}, we use pairs of economic status and statement of reason(s) from the Economy Watchers Survey as training data. The five-point-scale statuses $\{\circledcirc$, {\scriptsize$\bigcirc$}, \scalebox{0.8}{$\square$}, $\vartriangle$, $\times\}$ are converted to numerical values $\{2, 1, 0, -1, -2\}$, respectively, so as to treat it as a regression problem.

Two models are trained using the training data; One is a model to predict an economic status as a continuous value for an input text, and the other is an outlier detection model described in the next section. Regarding the former, we adopt a language representation model, specifically, BERT~\citep{devlin19:_bert}, for its superior performance to existing models.  BERT and its derivative models are based on Transformers~\citep{vaswani17:_atten_all_you_need} and have been widely used in recent years.  These models are initially learned on a large-scale, unlabeled corpus by solving a task to predict masked words from their context.  The initial models can be fine-tuned for various downstream tasks by feeding a task-specific labeled corpus and it has been reported that they yield superior performance to the former state-of-the-art models. We use a pre-trained Japanese BERT model\footnote{\url{https://github.com/cl-tohoku/bert-japanese}} and add an output layer that predicts economic status or a \textit{business sentiment score} of an arbitrary input text.  The weights of the entire model are fine-tuned using the Economy Watchers Survey as training data.  Our work is one of the earliest attempts, if not the first, to use a language representation model to predict business sentiment.

\subsubsection{Outlier Detection Model}
\label{sec:outlier}

One could use the aforementioned fine-tuned BERT model to predict a business sentiment score for any input text. However, news articles are generally written in many genres, some of which may not be necessarily relevant to the economy.  Using irrelevant information would be harmful in capturing business sentiment.  Therefore, we attempt to filter out such irrelevant news texts by treating them as outliers.  For this purpose, we preliminarily compared several outlier detection models and chose a one-class support vector machine (SVM)~\citep{manevitz02:_one_svms_docum_class}.  Unlike an ordinal SVM used for binary classification, a one-class SVM can be learned on documents belonging to only one class and detect documents dissimilar to the training documents as outliers.  We use Economy Watchers Survey, specifically, statements of the reason(s), as the training data for one-class SVM and filter out news text dissimilar to those statements.  For text representation, we use the traditional Bag-of-Words (BoW) with the term frequency-inverted document frequency (tfidf) term weighting~\citep{Manning:2008:IIR:1394399}.  In Section~\ref{sec:outlier-detection}, we will empirically compare one-class SVM with an alternative outlier detection model to discuss its advantage.  Note that there are several prior works to predict business sentiment from textual data as summarized in Section~\ref{sec:related_work} but few paid attention to whether or not each input text should be used for predicting business sentiment.  In contrast, we use an outlier detection model to selectively use input sentences from news articles.

\subsection{S-APIR Computation}
\label{sec:regression}

For computing the S-APIR index, we take advantage of newspaper articles.  First, each news article is divided into sentences based on the Japanese punctuation mark ``\Ja{。}$\!\!\!$'' and fed to the outlier detection model.  The sentences judged as outliers are filtered out and the other sentences are input to the fine-tuned BERT model. As a result, a business sentiment score is obtained for each sentence. The output scores can be aggregated by any arbitrary unit, e.g., daily or weekly.  Following the related work~\citep{aiba18eng,goshima19eng,kondo19eng}, we aggregate them by their average to define the S-APIR index and use monthly S-APIR throughout this paper as it can be directly compared with EWDI.

\subsection{Word-Level Temporal Analysis}
\label{sec:contribution}

Business sentiment is formed by various factors including monetary policies, trade, military conflicts, and an outbreak of pandemic diseases.  However, it is not clear how each factor influences the overall sentiment and discovering the influence of those potential factors has a tremendous value.  For instance, policymakers could assess the impact of any event of their interest (e.g., certain fiscal policies) and take necessary measures in a timely fashion.

To this end, we propose a simple approach to analyzing \textit{when} and \textit{how much} any given factor contributed to the business sentiment index.  Specifically, we define the contribution of word $w$ during time $t$, denoted as $p_{t,w}$, using the predicted business sentiment score of an input news text.  We first assume that the sentiment score $p_s$ of sentence $s$ is additive, that is, $p_s$ is the sum of the sentiments of words ($w$) appearing in $s$ as follows: \begin{equation}
  p_s=\sum_{i=1}^{N_s} p_{s,w_i} 
\end{equation}
where $N_s$ is the number of words composing $s$ and $p_{s,w}$ is the sentiment of $w$.  Further assume that all the words $w_i$ ($i=1,\ldots,N_s$) equally contribute to the sentiment of $s$. Then, $p_{s,w_i}$ is simply written as:
\begin{equation}
  p_{s,w_i}=\frac{p_s}{N_s}\ .
  \label{eq:p_sw}
\end{equation}
Note that $p_s$ is the output of the fine-tuned BERT model for input sentence $s$.  Here, let $S_t$ denote the set of news sentences published during time $t$. Using $S_t$, we define the contribution of $w$ in time $t$ as the average of $p_{s,w}$ over $S_t$.
\begin{eqnarray}
  p_{t,w}=\frac{1}{|S_t|}\sum_{s\in S_t} \frac{p_s}{N_s}
\end{eqnarray}
In cases where $w$ is a compound word, we multiply Equation~(\ref{eq:p_sw}) by the number of constituents of $w$. Intuitively, the S-APIR index in time $t$ can be interpreted as the sum of the influences of all the words appearing in texts published during $t$.  To the best of our knowledge, there has never been an attempt to temporally analyze business sentiment at a word level.

\section{Evaluation}
\label{sec:evlaluation}

\subsection{Experimental Settings}
\label{sec:data}

For fine-tuning BERT and a one-class SVM, we downloaded the Economy Watchers Survey data from the web page of the Cabinet Office\footnote{\url{http://www5.cao.go.jp/keizai3/watcher/watcher_menu.html}} in February 2020.  The number of the pairs of an economic condition and a statement of the reason(s) was 254,823 in total, of which randomly selected 90\% were used for training and validation and the rest were used for testing.  The ratio of the training and validation data was set to 9:1.

When fine-tuning a BERT model, we tested a number of combinations of parameters and set the batch size and the number of epochs to 32 and 3, respectively, which resulted in the least mean squared error (MSE) on the validation data.  The length of the input word sequence was set to 200.  To compute the S-APIR index, we used the titles and body texts of news articles from the Nikkei newspaper\footnote{The Nikkei   (formally the Nihon Keizai Shimbun) is a national financial newspaper in Japan.}  from January 2008 to June 2020.  The titles and body texts were not distinguished.  

\subsection{Preliminary Experiment}
\label{sec:keiki_watcher_index}

As a preliminary experiment, we evaluated the fine-tuned BERT on the held-out test data (10\% of Economy Watchers Survey).  In other words, the model was tested how closely it could predict the economic condition for a given statement of the reason(s). Table~\ref{tab:regression} compares our model and two other models, ridge regression and LSTM-BiRNN~\citep{yamamoto16eng}, in MSE.  For ridge regression, we used the BoW representation with tfidf term weighting.

\begin{table}[htb]
  \small
  \centering
  \caption{Model comparison in MSE for predicting economic
    conditions.}
  \label{tab:regression}
  \smallskip
  \begin{tabular}{cc}
    \hline
    Model & MSE \\
    \hline
    Ridge regression & 0.494\\
    LSTM-BiRNN~\citep{yamamoto16eng}  & 0.355\\
    BERT (ours) & 0.321\\
    \hline
  \end{tabular}
\end{table}

As compared to ridge regression with BoW which disregards the context of words and LSTM-BiRNN used in the related work~\citep{yamamoto16eng}, the fine-tuned BERT yielded the smallest MSE.  This result confirms the advantage of transformer-based language representation models over previous models as witnessed in other downstream NLP tasks~\citep{devlin19:_bert}.

\subsection{Evaluation of S-APIR}
\label{sec:nikkei_index}

This section compares our business sentiment index, S-APIR, and existing business sentiment index, specifically EWDI.  It should be emphasized, however, that S-APIR is intended not to replace EWDI but to be a new index using newspaper as the source of information.  There is no ground truth for a business sentiment index and EWDI is also one of the possible indices, which is measured based on somewhat limited 2,050 respondents.  Nevertheless, we make the comparison to ensure that S-APIR generally has a similar trend to the existing index and to study the characteristics of S-APIR.

\begin{figure}[tb]
  \centering
  \includegraphics[width=\linewidth]{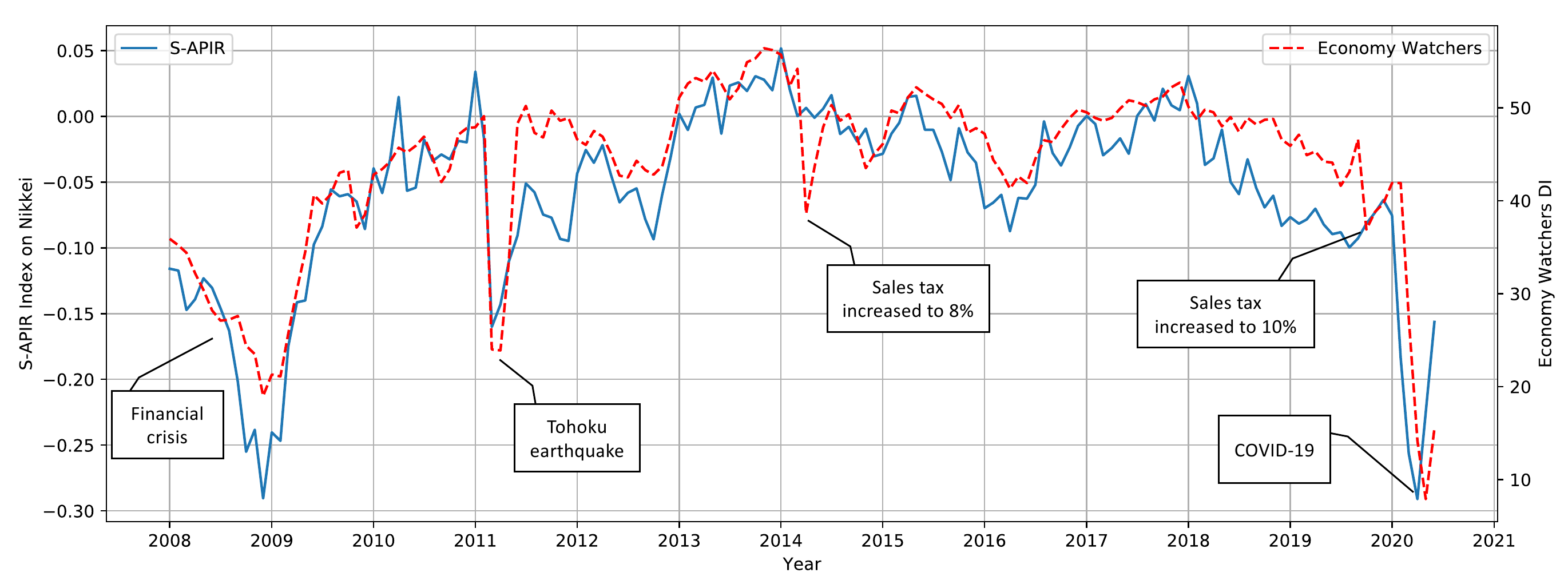}
  \caption{Comparison between S-APIR and EWDI ($r=0.888$).}
  \label{fig:nikkei}
\end{figure}

Using the fine-tuned BERT and the one-class SVM as described in Section~\ref{sec:keiki_watcher_index}, we computed monthly S-APIR on the Nikkei Newspaper from January 2008 to June 2020. Figure~\ref{fig:nikkei} shows the computed S-APIR and compares it with EWDI.  One can observe that S-APIR's trend is generally close to that of EWDI, capturing the financial crisis triggered by the bankruptcy of Lehman Brothers in 2008 and the decline of business sentiment caused by the Great East Japan Earthquake (Tohoku Earthquake) in 2011. In effect, they were found to be strongly positively correlated ($r=0.888$), which proves the validity of the S-APIR index.

Here, we would like to remind that the input texts used in this experiment are only newspaper articles.  The result is striking considering the fact that EWDI is calculated based on a costly survey specially designed to measure business sentiment, whereas the Nikkei is a national newspaper for the general public even though it focuses on economy and business.  While we do not have complete understanding of why S-APIR has strong positive correlation with EWDI, one of the reasons would be that the prediction model was trained on EWDI's survey data.  Another reason would be that the Nikkei Newspaper contains many articles in economy, finance, and business and consequently its contents are similar to some extent to EWDI's survey responses, which are mainly the situations of economy and business the respondents observed.  

Incidentally, Economy Watchers Survey contains occupations of respondents as shown in Table~\ref{tab:keiki_watcher}, which are categorized into household-related (about 70\% of the respondents), industry-related (about 20\%), and employment-related (about 10\%). Thus, the influence of household trends is relatively large on EWDI.  On the other hand, the Nikkei used in this study is a financial newspaper publishing many business-related articles.  Consequently, S-APIR is likely to be more influenced by businesses and industries.  To examine the assumption, we compared S-APIR with industry-related EWDI which was computed based only on industry-related respondents.  Then, their correlation coefficient increased from 0.888 to 0.937. This result indicates the characteristics of the S-APIR index calculated from the Nikkei that it reflects industry trends more closely.

\subsection{Outlier detection}
\label{sec:outlier-detection}

We used one-class SVM to filter out sentences irrelevant to the economy in Section~\ref{sec:nikkei_index}.  In order to validate the effectiveness of the filtering process, we computed an S-APIR index based on all the news sentences without applying filtering.  As a result, the correlation coefficient with EWDI slightly decreased from 0.888 to 0.878.  When compared with industry-related EWDI, the difference was somewhat greater; it decreased from 0.937 to 0.919.  In both cases, correlation coefficients decreased, suggesting that news articles contain sentences not suitable to measure business sentiment and that one-class SVM was able to filter them out.

We also tested the effectiveness of the filtering process when LSTM-BiRNN (see Section~\ref{sec:keiki_watcher_index}) was used for computing a business sentiment index.  The correlation coefficient with EWDI was found to be 0.765 without filtering and 0.875 with filtering.  The correlation efficient with industry-related EWDI was 0.805 without filtering and 0.922 with filtering.  In both cases, the benefit of the filtering was much greater than that of BERT.  The results indicate that BERT is more robust than LSTM-BiRNN when input texts contain sentences irrelevant to economy.

Furthermore, to examine the validity of the use of one-class SVM, we tested an alternative outlier detection model based on an LSTM-RNN auto-encoder~\citep{8411269}. The auto-encoder uses LSTM units in a hidden layer of RNN and judges an input sequence as an outlier if the reconstruction error is greater than a predefined threshold.  For training the model, 90\% of the Economy Watchers data (see Section~\ref{sec:data}) were used and the remaining 10\% were used as validation data.  As model parameters, we explored the combinations of the dimension of word embedding vectors $d \in \{100, 200, 300\}$, the number of units $u \in \{32, 64\}$ in a hidden layer, and batch size $b \in \{8, 16\}$ by grid search.  The best model with the least loss on the validation data was obtained with $d=300$, $u=64$, and $b=16$.  We did not test larger parameter values due to the limitation of our computing resources.

In order to directly compare the one-class SVM and LSTM-RNN autoencoder models for outlier detection, a data set consisting of inliers and outliers is needed. However, it is costly to manually create such a dataset. Instead, we used as inliers 7,962 data from the Economy Watchers Survey after March 2020, which were not used for training the outlier detection models. As for outliers, we used 14,912 articles in the entertainment category in the 2019 edition of the Mainichi Newspaper.\footnote{The Mainichi Newspaper is one of the four national newspapers in Japan.} Although there may be articles related to the economy (i.e., inliers) even in the entertainment category, it is expected that the majority would be outliers. The results are shown in Table~\ref{tab:outlier}.  Recall, precision, and $F_1$ scores were macro-averaged for the inliers and outliers classes. Note that, for autoencoder, it is necessary to set a threshold for the reconstruction error to judge whether it is an outlier or not.  We tested a number of threshold values and Table~\ref{tab:outlier} shows the performance with the highest $F_1$.

\begin{table}[tb]
  \centering
  \caption{Comparison between one-class SVM and an alternative outlier
    detection model based on LSTM-RNN autoencoder.}
  \label{tab:outlier}
  \smallskip
  \begin{tabular}{@{~}lccc@{~}}
    \hline 
    \multicolumn{1}{c}{Model} & Recall & Precision & $F_1$ \\
    \hline
    auto-encoder & 0.928 & 0.858 & 0.892\\
    one-class SVM & 0.929 & 0.916 & 0.923 \\
    \hline 
  \end{tabular} 
\end{table}

From the result, it can be seen that the one-class SVM has greater performance for outlier detection. Intuitively, an RNN-based autoencoder has an advantage as it can capture contextual information.  However, for this relatively simple task to identify whether a sentence is about the economy, the word-level features used by one-class SVM appear to suffice. On trial, we used the auto-encoder for outlier detection and computed the business sentiment index from January 2008 to June 2020.  As a result, the correlation coefficient with EWDI was found to be 0.875, which is slightly worse than the case where outlier detection was not used (0.878).

\subsection{Business Sentiment by General Newspaper}
\label{sec:general-newspaper}

We computed the S-APIR index based on the Nikkei newspaper as it focuses on economics, finance, and business and was deemed suitable for measuring business sentiment.  The experiment presented in Section~\ref{sec:nikkei_index} supported the expectation, showing that the S-APIR index has a strong correlation, especially with the industry-related EWDI.

To further investigate the relation between input text and the resulting business sentiment index, we measured the business sentiment using general newspaper as input and compared it with Nikkei's result. Specifically, we calculated an S-APIR index using Mainichi newspaper articles from January 2008 to December 2019 with/without filtering (outlier detection). The results are summarized in Table~\ref{tab:mainichi}.  In the table, Nikkei's result is also shown for comparison, which is slightly different from Section~\ref{sec:nikkei_index} because the articles from January to June 2020 were excluded from the analysis in accordance with the Mainichi newspaper used for this experiment.

\begin{table}[tb]
  \centering
  \caption{Correlation coefficient between S-APIR and EWDI when
    different newspaper were used as input. The figures in parentheses
    indicate percent increase when filtering was applied.}
  \label{tab:mainichi}
  \smallskip
  \begin{tabular}{ccr@{~}l}
    \hline 
    Source & Filtering & \multicolumn{2}{c}{Correlation}\\
    \hline
    Nikkei  & no & 0.873 &\\ 
     & yes & 0.892 & (+2.1\%)\\ 
    Mainichi  & no & 0.738 & \\
      & yes & 0.817 & (+10.7\%)\\
    \hline 
  \end{tabular} 
\end{table}

From this result, we can make two observations. Firstly, the Nikkei newspaper, as expected, is more useful for predicting business sentiment. Secondly, the effect of filtering is greater for Mainichi newspaper. Specifically, Nikkei showed a 2.1\% improvement in the correlation coefficient due to filtering, while the Mainichi showed a 10.7\% improvement. This result is consistent with the intuition that Mainichi, which is a general newspaper, contains fewer articles related to the economy and business as a whole.

Furthermore, as an attempt, we computed the S-APIR index using both Nikkei and Mainichi, and the correlation coefficient was slightly improved to 0.899 as compared to using either newspaper alone.  The result is interesting from the viewpoint of the use of big data that, even though Mainichi is suboptimal as compared to Nikkei, their combination works complementarily.

\subsection{Relationship between S-APIR and Business Conditions
  Indicators}
\label{sec:alternatives}

This section looks at the relationship between the S-APIR index and other business conditions indicators. S-APIR is computed based on a fine-tuned BERT model, which was learned from the Economic Watchers data. Consequently, S-APIR was found to be strongly correlated with EWDI even though S-APIR was computed \textit{not} from survey responses \textit{but} from newspaper articles.  While the strong correlation with EWDI is beneficial on its own right, EWDI is just one economic indicator among others based on a survey.  Therefore, it is crucial to study the characteristics of S-APIR as a more general business conditions indicator in comparison with other representative ones.

\begin{figure}[tb]
  \centering
  \includegraphics[width=\linewidth]{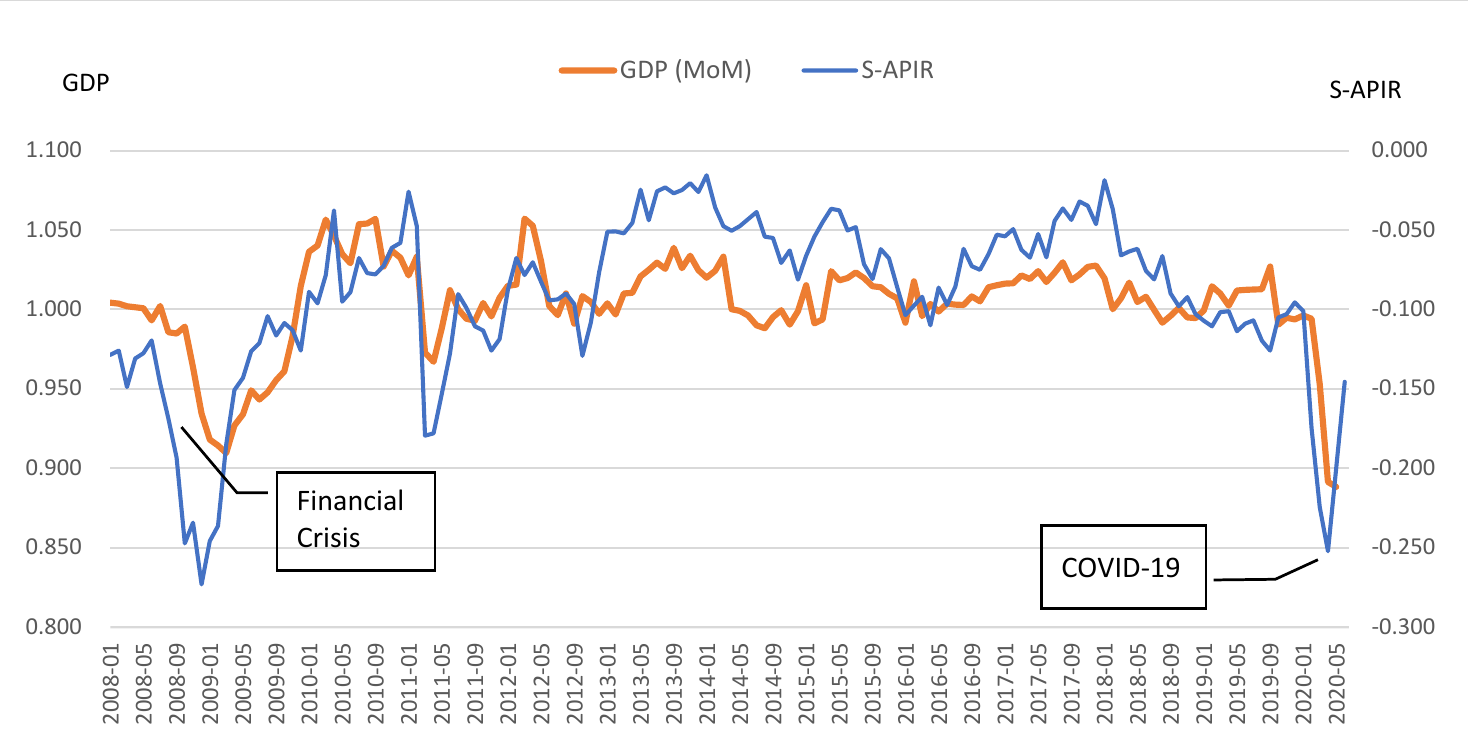}
  \caption{Comparison between S-APIR and MoM GDP ($r=0.698$).}
  \label{fig:GDP}
\end{figure}

First, we examined the relationship with two macroeconomic indicators. The first is the Gross Domestic Product (GDP) of Japan. Figure~\ref{fig:GDP} compares the S-APIR index and month-on-month GDP. As can be seen, S-APIR generally followed the movement of GDP and their correlation coefficient was 0.698.

Looking into the details, S-APIR is slightly ahead of GDP when there are major shocks. For example, during the financial crisis after the collapse of Lehman Brothers in 2008, GDP began to decline around the end of the year and bottomed out in early 2009. S-APIR, on the other hand, already started to decline rapidly around September 2008 and bottomed out in early 2009. For almost half a year, S-APIR had been ahead of the economic trend. The S-APIR is also slightly ahead of the recession by the COVID-19 pandemic in 2020.

\begin{figure}[tb]
  \centering
  \includegraphics[width=.7\linewidth]{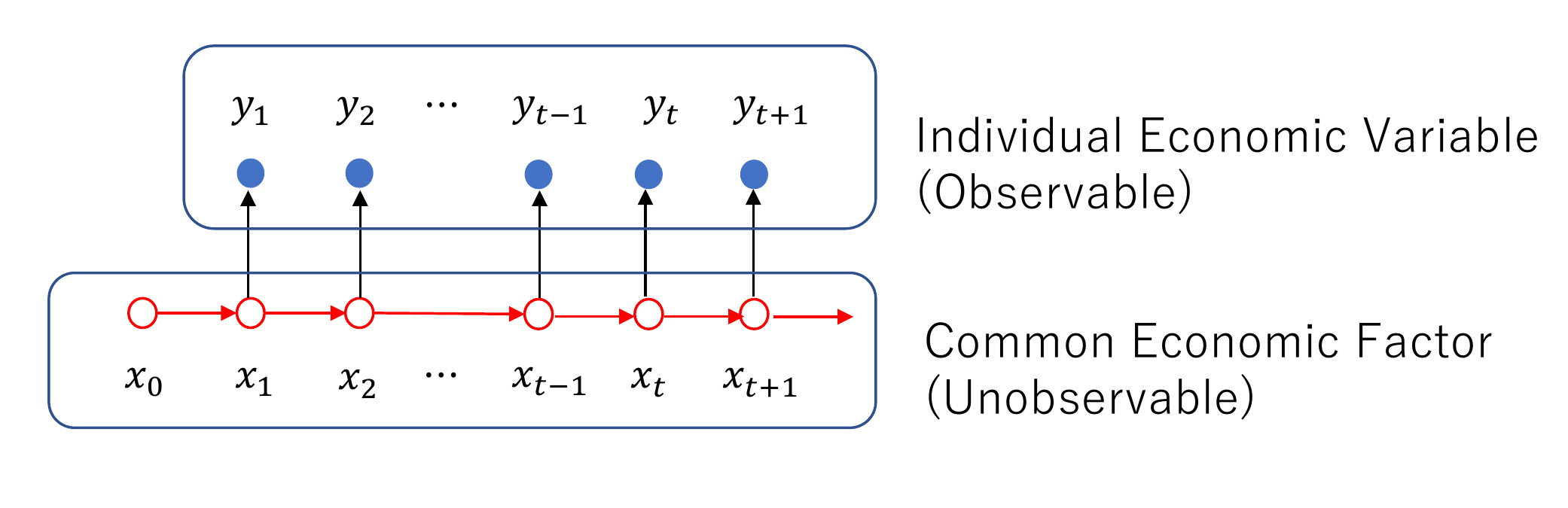}
  \caption{An illustration of the Dynamic Factor Model (DFM).}
  \label{fig:DFM_img}
\end{figure}

The second macroeconomic indicator to be compared is an artificially measured one. Specifically, we used the dynamic factor model (DFM)~\citep{stock1989,stock1991} to measure the common factors to several representative economic indicators using a state-space model.\footnote{The business condition index by DFM was developed by \cite{stock1989,stock1991} as a Stock-Watson type index. The estimates are published by the National Bureau of Economic Research (NBER).}  Figure~\ref{fig:DFM_img} illustrates DFM, which assumes that observable individual economic data $y_{i,t}$, such as Indices of Industrial Production (IIP), are generated from unobservable common macroeconomic indicator $x_t$, where $i\in\{1,\ldots,N\}$ denotes an individual economic index and $t\in\{1,\ldots,T\}$ denotes a time-series index.

The relationship between $y_{i,t}$ and $x_t$ are formulated by the single-index DFM as follows:
\begin{eqnarray}
\label{measurement}
  y_{i,t} & = &  \beta_{i,0} + \gamma_i x_t + u_{i,t} \\
\label{transition} 
  x_t & = & \phi_1 x_{t-1} + \phi_2 x_{t-2} + \cdots + \phi_p x_{t-p} + \eta_t  \\
\label{shock}
  u_{i,t} & = & d_{i,1} u_{i,t-1} + d_{i,2} u_{i,t-2} + \cdots + d_{i,q} u_{i,t-q} + \epsilon_{i,t} 
\end{eqnarray}
where $u_{i,t}$ is an idiosyncratic shock which is not correlated with $x_t$ identically, and $\eta_t$ and $\epsilon_{i,t}$ are error terms.  

Equation (\ref{measurement}) is a measurement equation, where individual economic variable $ y_{i,t} $ depends on $ x_t $ and $ u_{i,t}$. Equation (\ref{transition}) is a transition equation, where $ x_t $ follows a $p$-order autoregressive process. An idiosyncratic shock $ u_{i,t} $ follows a $q$-order autoregressive process as in Equation (\ref{shock}).  The equations are transformed to state-space representation since $x_t$ is not observable. Then, parameters determining the relationship between $y_{i,t}$ and $x_t$ are estimated by Kalman filtering.

In this study, we used as $y_{i,t}$ four types of data regarding production, consumption, employment, and exports (i.e., $N=4$). These four observable economic variables are assumed to be basic series to estimate the common, unobservable economic indicator $x_t$.

\begin{figure}[tb]
  \centering
  \includegraphics[width=\linewidth]{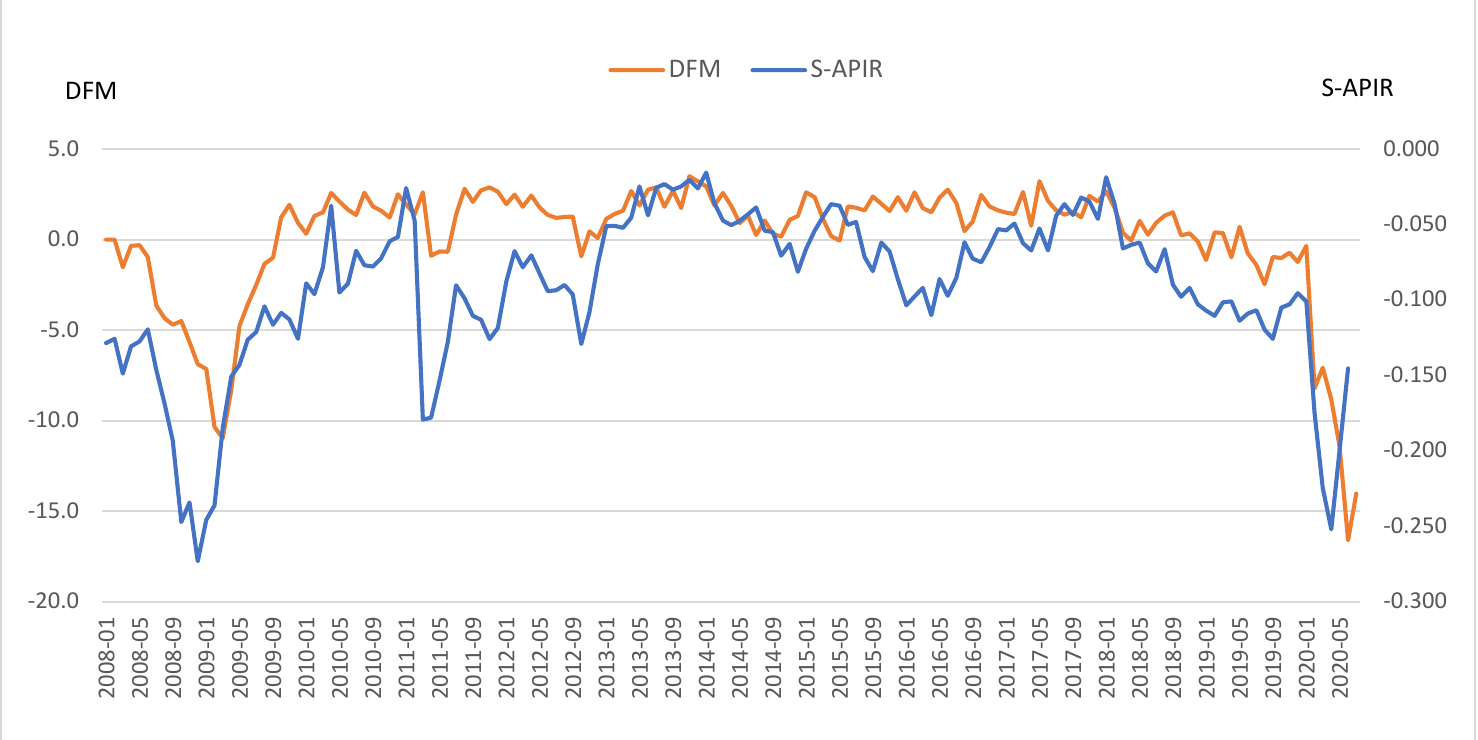}
  \caption{Comparison between S-APIR and DFM ($r=0.749$).}
  \label{fig:DFM}
\end{figure}

As shown in Figure~\ref{fig:DFM}, S-APIR and DFM have roughly similar movements ($r=0.749$).  Also, as in the case of GDP, S-APIR shows a slightly earlier movement than DFM for major shocks.

As can be seen from the above observations, the S-APIR index is similar to the GDP and DFM in terms of its overall trends, indicating its characteristic that it is able to capture the macroeconomy as a whole. Also, the S-APIR is ahead of the macroeconomic indicators to some extent, suggesting another characteristic that it encompasses some information concerning the economic outlook and sentiment of economic agents (e.g., consumers and producers).

To analyze the latter characteristic in more detail, we examine the relationship between S-APIR and several semi-macro indicators. For sentiment indices, we focus on the Consumer Confidence Index and the Purchasing Managers' Index (PMI), which represent the aspects of consumers and sellers, respectively. For actual activity indices, we focus on the Synthetic Consumption Index computed from both demand- and supply-side statistics and the Consumer Activity Index computed from only the supply-side statistics.

\begin{figure}[tb]
  \centering
  \includegraphics[width=\linewidth]{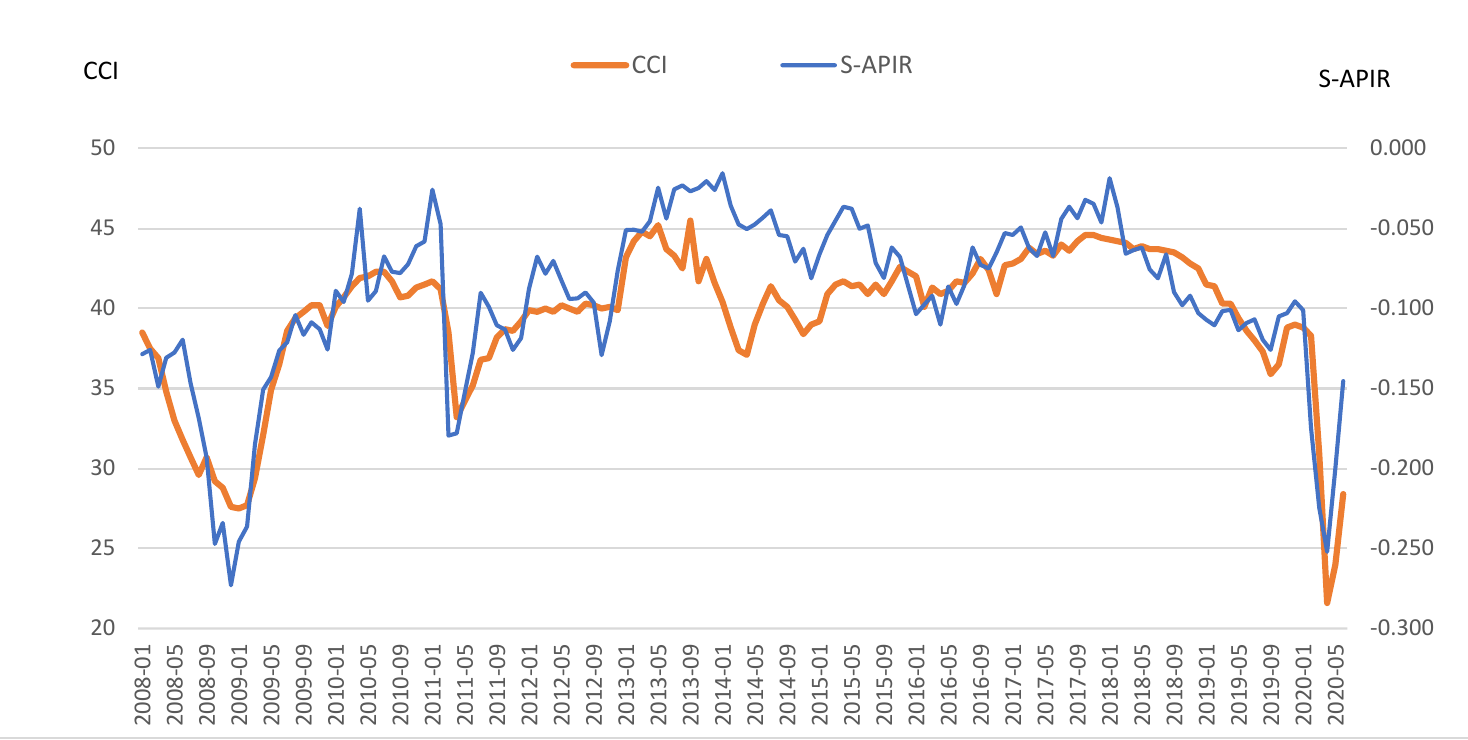}
  \caption{Comparison between S-APIR and Consumer Confidence Index
    (CCI) ($r=0.848$).}
  \label{fig:CCI}
\end{figure}

Consumer Confidence Index (CCI) is the most famous index of consumer confidence in Japan. Figure~\ref{fig:CCI} compares CCI with S-APIR, where their movements are similar, resulting in a strong positive correlation ($r=0.848$). In fact, the timing of the decline and recovery during major shocks, specifically, the financial crisis in 2008 and the COVID-19 pandemic in 2020, is generally consistent, suggesting that S-APIR reflects the consumer sentiment toward the future.

Next, we looked at the Manufacturing PMI (referred to as MPMI) and the Services PMI (referred to as SPMI) as outlook indicators from the sales side\footnote{The intellectual property rights to these data are   owned by IHS Markit.}.  The PMIs are calculated based on the results of monthly surveys obtained from companies' purchasing staff. The survey contains responses, including production, new orders, inventory level, employment status, and prices. Typically, a company's buyer purchases goods to meet demand and prepares for sale. For this reason, the PMIs look at economic activity from the standpoint of sales, which is ahead of consumption. Thus, PMIs can be seen as leading indicators of the economy through buyers who are sensitive to market movements. Since MPMI targets the manufacturing industry, it contains the information related to business-to-business (B2B) transactions, whereas SPMI targets the service industry, containing the information related to business-to-consumer (B2C) transactions.

\begin{figure}[tbp]
  \centering
  \includegraphics[width=\linewidth]{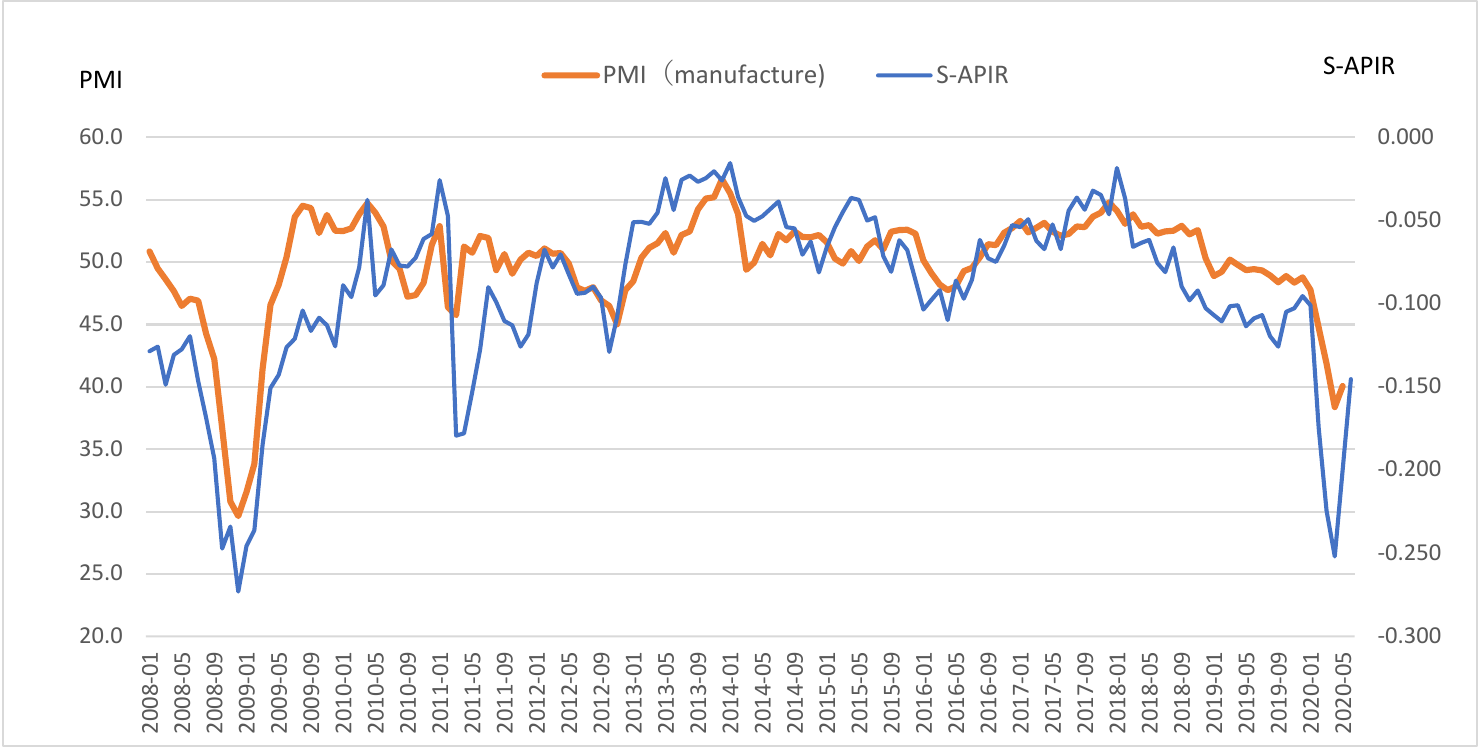}\\
  \includegraphics[width=\linewidth]{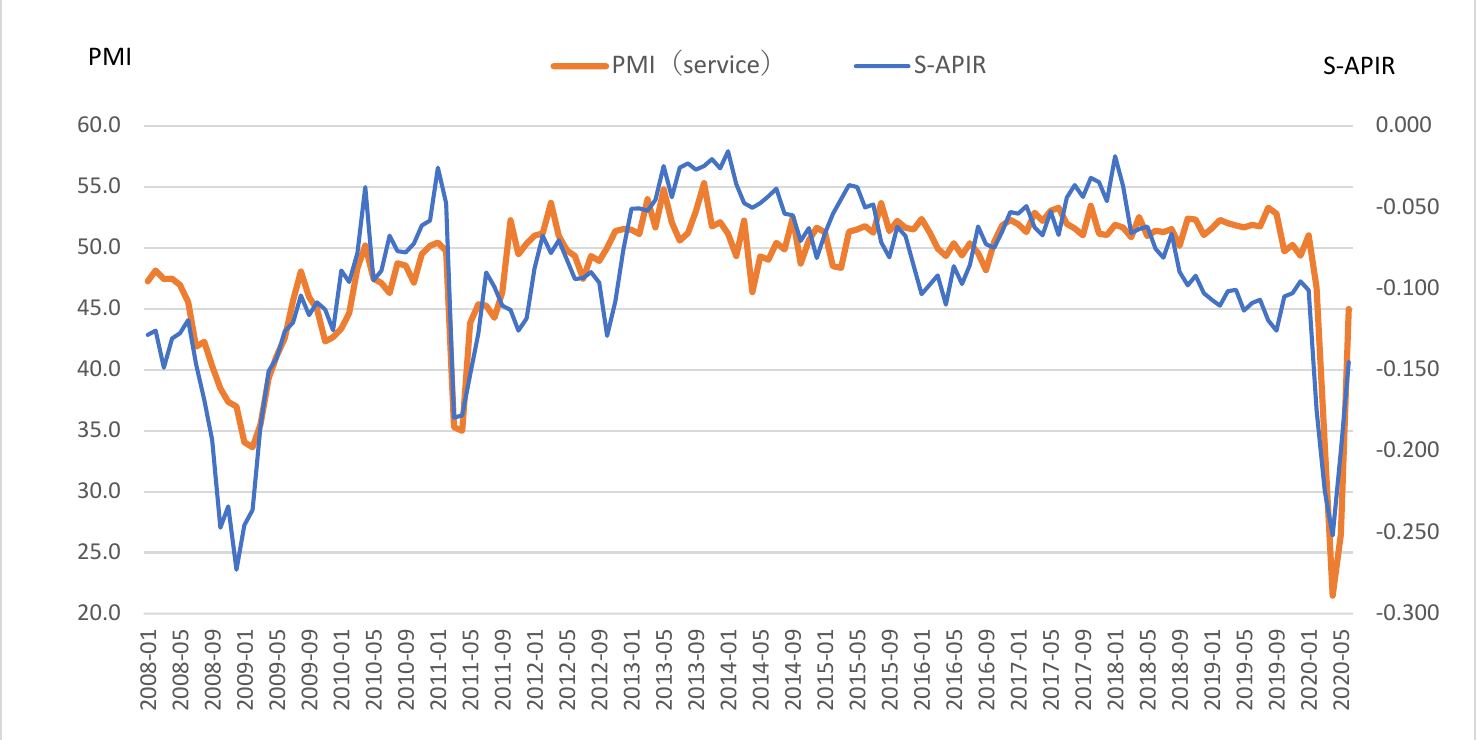}
  \caption{Comparison between S-APIR and MPMI ($r=0.773$) (top) and
    S-APIR and SPMI ($r=0.802$) (bottom).}
  \label{fig:PMI}
\end{figure}

Figure~\ref{fig:PMI} compares the PMIs with S-APIR.  We can observe that S-APIR more closely follows MPMI during the financial crisis in 2008 than SPMI.  On the other hand, S-APIR more closely follows SPMI during the COVID-19 pandemic in 2020 than MPMI.  The rapid drop of MPMI during the global financial crisis was partly due to the significant decrease in demand for automobiles. SPMI declined during the COVID-19 pandemic because the demand for food, retail, and accommodation has fallen sharply. S-APIR accurately captured the outlook of the two different types of sellers (i.e., B2B and B2C) during these periods, implying the coverage of Nikkei newspaper---the information source of the S-APIR index---is higher on the manufacturing or service industry, whichever affected worse, depending on a particular period.

\begin{figure}[tbp]
  \centering
  \includegraphics[width=\linewidth]{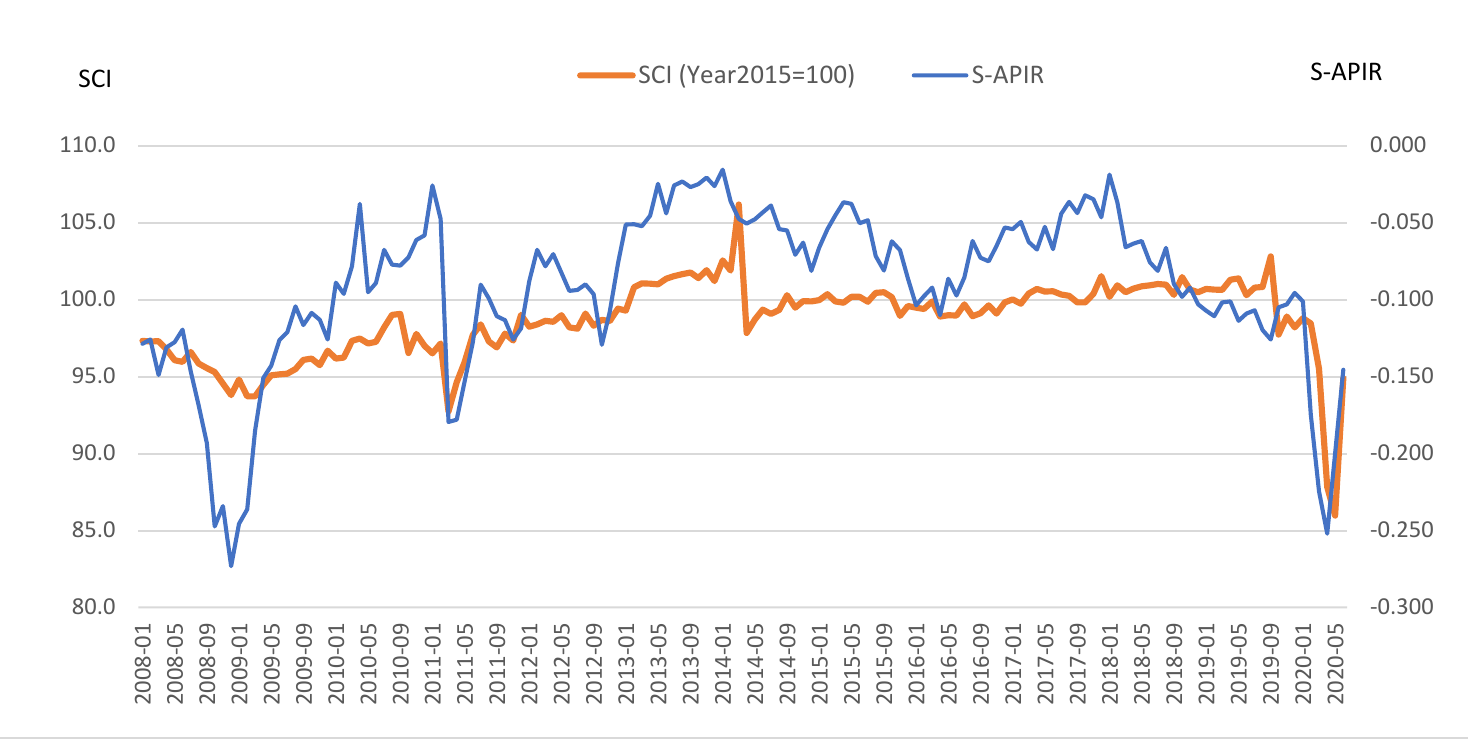}\\
  \includegraphics[width=\linewidth]{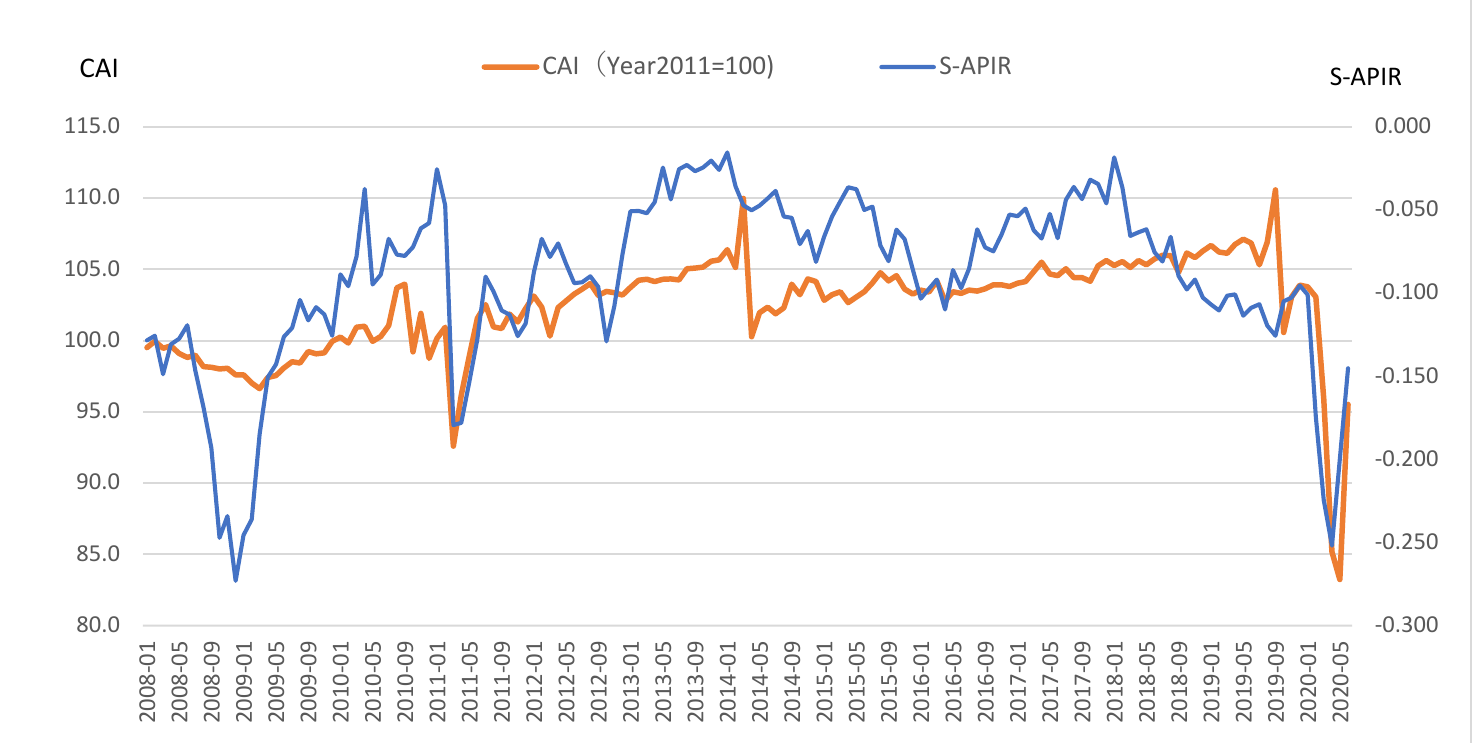}
  \caption{Comparison between S-APIR and SCI ($r=0.727$) (top) and
    S-APIR and CAI ($r=0.650$) (bottom).}
  \label{fig:CTI_CAI}
\end{figure}

The above comparison of S-APIR with sentiment indices (i.e., CCI and PMIs) showed that S-APIR well captured consumer activity forecasts. However, the result does not tell whether S-APIR precedes actual consumption activities. Therefore, by comparing S-APIR with actual activity indices (i.e., SCI and CAI), we study the suitability of S-APIR as a leading index of consumption activity.  In Figure~\ref{fig:CTI_CAI}, it can be seen that S-APIR is ahead of the actual consumption, especially in large economic events, including the financial crisis in 2008, the tax increase in 2014 and 2019, and the COVID-19 pandemic.  An exception is the Great East Japan Earthquake, where S-APIR appears to move in line with SCI and CAI.

From the series of comparisons discussed above, we identified two major characteristics of the S-APIR index.  Firstly, S-APIR captures the overall macroeconomic trends.  Secondly, it contains much information concerning the outlook for consumption and sales.  This is presumably due to the fact that the Economic Watchers Survey (i.e., the training data of our model) includes a variety of sentences expressing the outlook for household consumption and corporate sales. Also, the Nikkei newspaper (i.e., the information source of the S-APIR) contains relatively a large number of articles on consumer sentiment and corporate sales forecasts.

\subsection{Contributions of Events to Business Sentiment}
\label{sec:contrib}

This section examines the contributions of several events to S-APIR as business sentiment. In conventional quantitative analysis, when analyzing the degree to which a certain event or factor affects economic trends, a time series of that factor (e.g., crude oil prices) and an indicator representing economic trends (e.g., GDP) are examined using regression analysis.  In contrast, we focus on the input news texts used for measuring business sentiment and take advantage of the context in which the term ``crude oil price'' appeared by way of the predicted business sentiment scores of the news sentences.  In other words, we attempt to examine the factors that cause economic fluctuations based on the behavioral patterns and future forecasts of economic agents that cannot be captured by quantitative data alone.

\begin{figure}[tb]
  \centering
  \includegraphics[width=\linewidth]{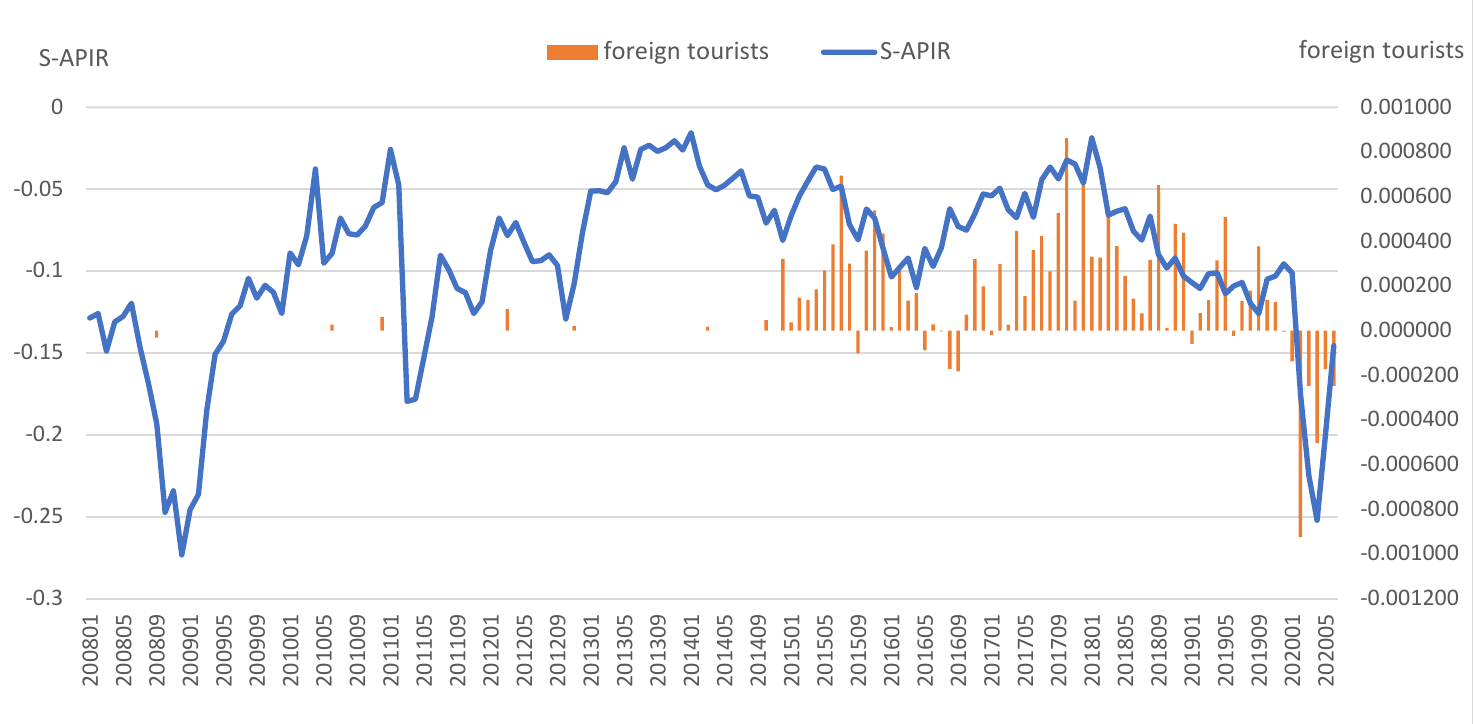}
  \caption{Contribution of ``\Ja{インバウンド}''
    (foreign tourists) to S-APIR index.}
  \label{fig:inbound}
\end{figure}

In the following, we focus on three events that have been attracting attention in the Japanese economy in recent years and examine their relationship with the S-APIR index. First, let us look at ``\Ja{インバウンド}'' (roughly translated to ``foreign tourists''). The term ``\Ja{インバウンド}'' is a loanword of ``inbound'' and the meaning was changed to refers to foreign tourists or foreign tourism to Japan, which had been increasing rapidly since the 2010s. As can be seen from Figure~\ref{fig:inbound}, the word began to have a positive effect on the economy around 2015 and it had been steadily rising until the end of 2017. In the mid-2010s, Japan saw the phenomenon of ``shopping spree'' by foreign tourists.  The plot confirms that this phenomenon lasted for about three years.

As mentioned earlier, our approach allows us to quantize how much and when various events affect economic sentiment in a timely manner. Since the beginning of the COVID-19 pandemic, foreign visits to Japan have been disrupted, which has made a great impact on the economy as shown in Figure~\ref{fig:inbound}.  At the same time, however, we can also observe that the impact is most severe in early 2020 and the situation has improved to some extent since then.  This suggests that the effect of the disruption of foreign tourism on the economy may not be very persistent, which is very useful for policymakers to assess the situation.  That is, if the slowdown in the economy due to the decline of foreign visitors is relatively transient, policymakers may not need to implement large-scale support measures for the tourism industry. Instead, they could allocate their limited resources to other industries and individuals needing immediate supports.

\begin{figure}[tb]
  \centering
  \includegraphics[width=\linewidth]{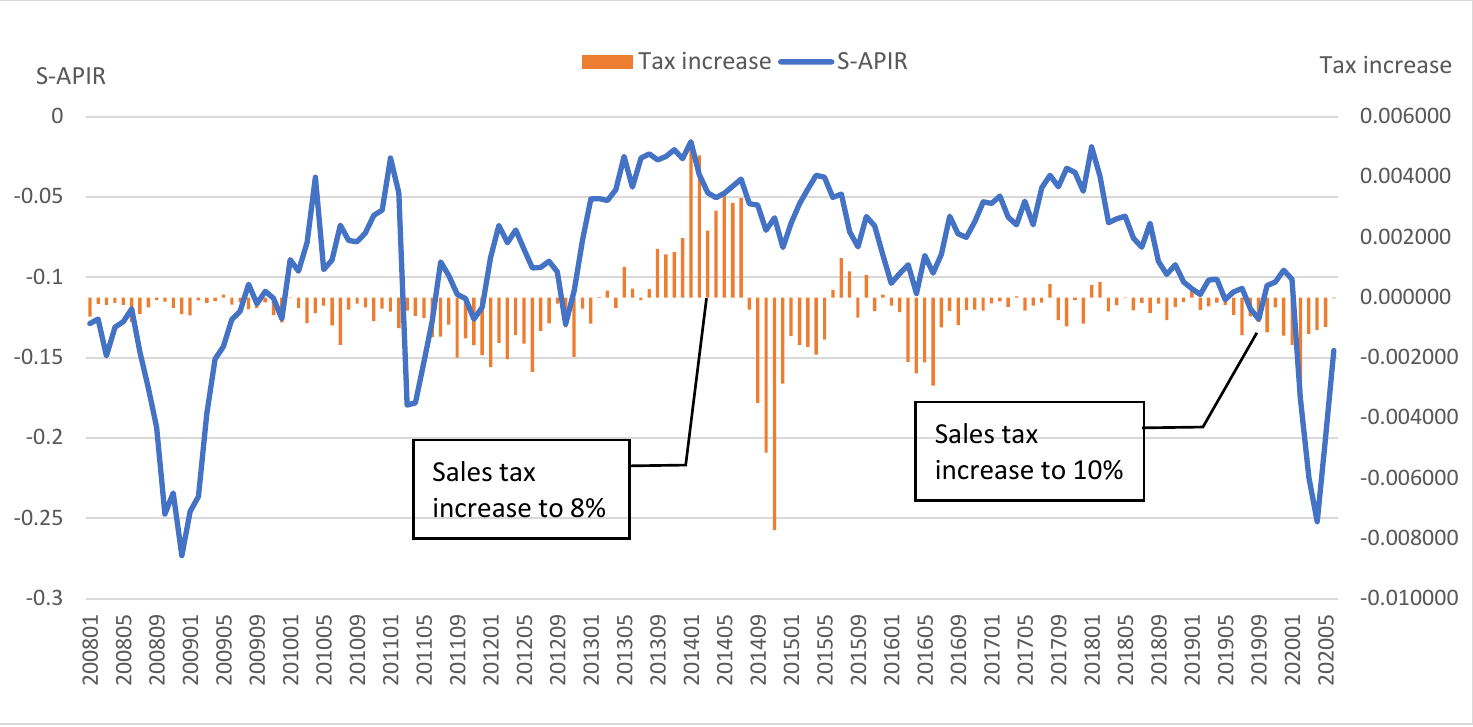}
  \caption{Contribution of ``tax increase'' to S-APIR.}
  \label{fig:taxincrease}
\end{figure}

Next, let us look at the relationship with ``\Ja{増税}'' (tax increase).  Since the 2010s, Japan had a consumption tax increase twice: the first increase from 5\% to 8\% took effect in April 2014 and the second from 8\% to 10\% in October 2019. Since the implementation of the consumption tax increase was announced in advance, it affected the economy through a rush of demand just before the tax increase and a decline afterward.  Figure~\ref{fig:taxincrease} clearly shows the effect of the first tax increase in 2014. However, when it was increased again in 2019, the impact of the rush demand was hardly seen and the rebound in consumption was subtle. It suggests that consumer sentiment has begun to lessen and is hardly responding to the shock of the tax increase.  The impact of the tax increase on the economic sentiment is also not something that can be easily confirmed by traditional quantitative analysis as in the case of foreign tourism to Japan, substantiating the utility of our approach.

\begin{figure}[tb]
  \centering
  \includegraphics[width=\linewidth]{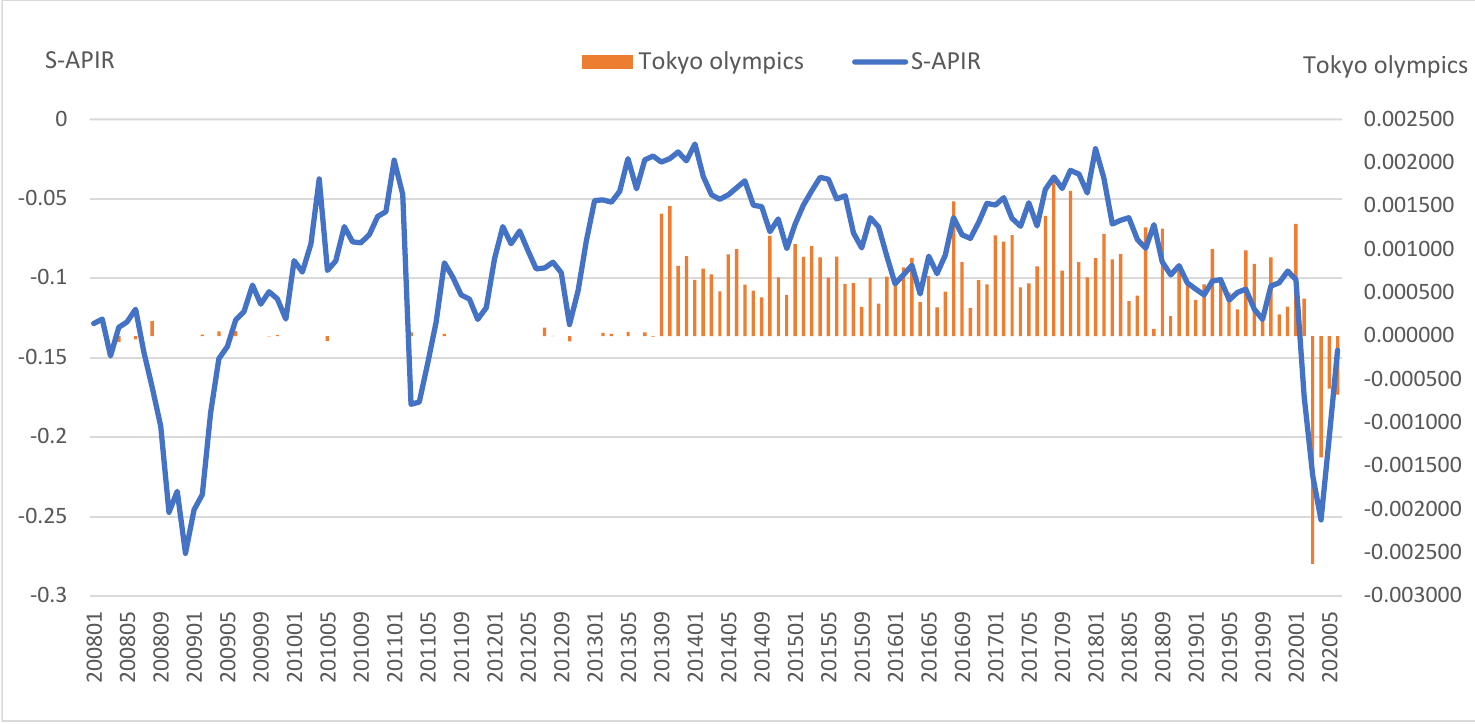}
  \caption{Contribution of ``Tokyo Olympics'' to S-APIR.}
  \label{fig:olympic}
\end{figure}

Finally, let us look at the major event, ``\Ja{東京五輪}'' (Tokyo Olympics), which was originally scheduled to be held in 2020.  As can be seen in Figure~\ref{fig:olympic}, the Olympics had generally a positive effect on the economy for about seven years from September 2013 when Tokyo was selected to host the Olympics.  However, as we moved into 2020, the impact on the economy has taken a turn for the worse as the COVID-19 pandemic has raised concerns about the event. We conjecture that, if the pandemic did not occur, we would have witnessed a constant and stable positive effect of the Olympics on the Japanese economy.  The impact of the Tokyo Olympics on the economy has taken nearly eight years since the selection of the host country, which makes it difficult for a traditional quantitative analysis to assess the impact of the event for a specific period on the economic sentiment.  In contrast, our approach enables a temporal analysis for any given event at any given time as long as it is covered by the newspaper.

These examples discussed above demonstrated how the effect of an event on business sentiment can be analyzed by our approach.  However, the approach is based on an assumption that all words contribute to the sentiment of a sentence, which can be argued as over-simplification. An alternative approach would be to take advantage of self-attention weights~\citep{vaswani17:_atten_all_you_need}.  BERT, or other attention-based language representation models, estimates attention weights for each element (word) of an input sequence in predicting the business sentiment score of the input. However, the values of attention weights tend to be similar as they approach the last attention layer and thus they do not necessarily represent the importance of words~\citep{serrano-smith-2019-attention}.  To represent their importance more properly, \cite{abnar20:_quant_atten_flow_trans} proposed attention flow and attention rollout.  The former takes attentions as a directed graph and computes the maximum flow, and the latter rolls out the attention weights to capture the propagation of information from input tokens to intermediate hidden embeddings.

We adopted attention rollout, which is computationally less expensive, and computed attention rollout $r_w$ from the \texttt{CLS} token (a special symbol representing the whole sentence) in the last attention layer of BERT to each word $w$ and distribute the sentiment $p_s$ of sentence $s$ to words $w$ proportionally to $r_w$.  That is, we used Equation~(\ref{eq:p_sw_att_flow}) instead of Equation~(\ref{eq:p_sw}) to estimate the sentiment $p_{s,w}$ of word $w$ in sentence $s$.
\begin{equation}
  p_{s,w}=p_s\cdot\frac{r_w}{\sum_{w' \in s}r_{w'}}
  \label{eq:p_sw_att_flow}
\end{equation}

\begin{figure}[tb]
  \centering
  \includegraphics[width=.8\linewidth]{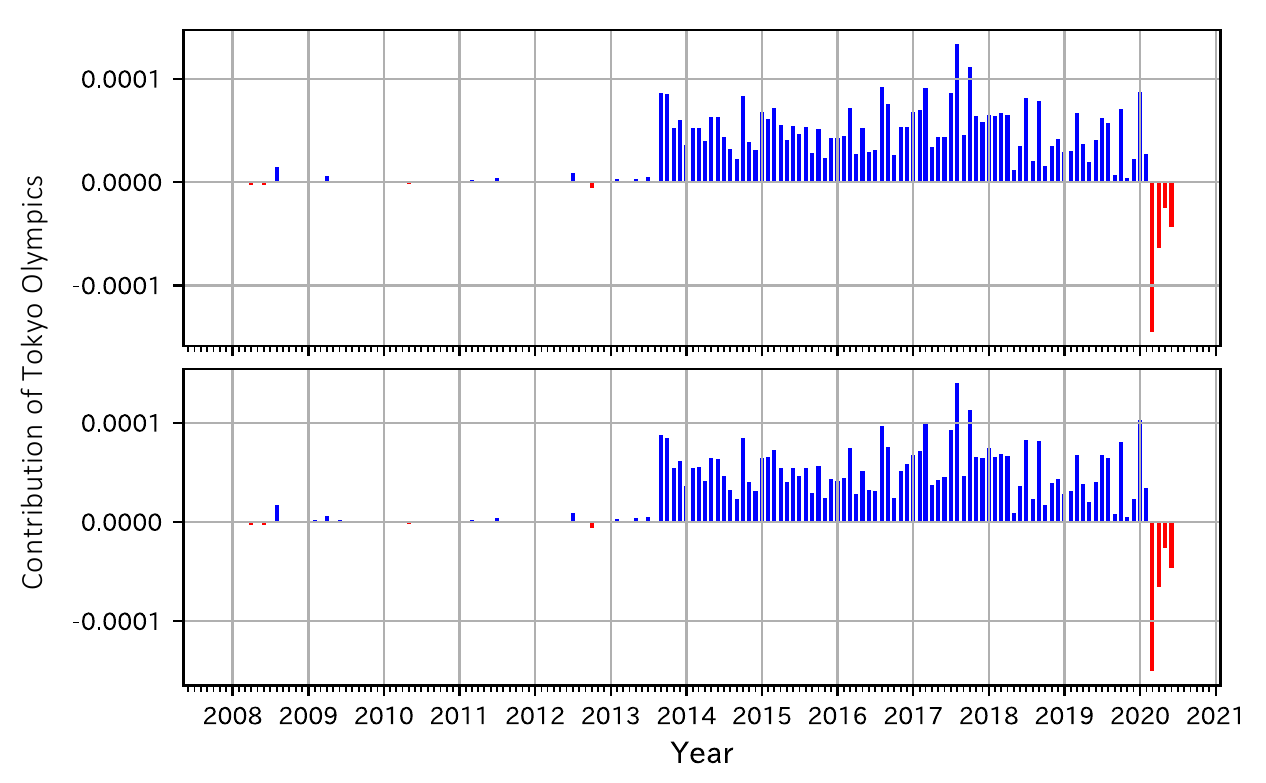}
  \caption{Contribution of Tokyo Olympics to S-APIR index, where a sentence sentiment was divided proportionally to attention rollouts for the upper figure and it was evenly distributed to its constituent words for the bottom figure.  }
  \label{fig:olympic_att}
\end{figure}

We recomputed the contribution of Tokyo Olympics over time using Equation~(\ref{eq:p_sw_att_flow}).  The result is shown in Figure~\ref{fig:olympic_att}, where the plot from Figure~\ref{fig:olympic} is also shown for comparison.  We can observe that they are almost identical with a few negligible differences (e.g., the contribution is slightly higher for the bottom plot than the top in January 2020).  The observations are similar for the other two events, tax increase and foreign tourism, and thus omitted.  It indicates that the assumption introduced in Equation~(\ref{eq:p_sw}) does not affect the result of the analysis much as compared to the cases where the importance of words was computed by attention weights.

\section{Discussion}
\label{sec:implications}

We presented an approach to turning news articles into a business sentiment index, S-APIR.  With the proposed index, we pursued three research objectives (RO) as stated in Section~\ref{sec:research-objectives}, which guided this work and distinguished it from the previous work.  Specifically, (a) we employed a self-attention-based language representation model, BERT, to measure business sentiment and used daily newspaper articles as input; (b) we explored effective outlier detection models for this particular problem; (c) we thoroughly investigated the properties of the S-APIR index by comparing it with a variety of economic indicators; and (d) we proposed a simple approach to temporally analyzing the influence of a given event on business sentiment.  The following discusses what we learned for each of the research objectives and summarizes the major findings.

\subsection*{RO1: To measure business sentiment based on daily newspaper articles and empirically validate the proposed approach.}
\label{sec:ro1}

\begin{itemize}

\item In a preliminary study, we fine-tuned a Japanese BERT model and compared it with the related work.  Our fine-tuned model outperformed other models including LSTM-BiRNN~\citep{yamamoto16eng} for predicting business conditions for a given statement of the reason(s) from the Economy Watchers Survey.

\item We fed the Nikkei newspaper published between 2008 and 2020 to the model to predict their sentiment scores.  The predicted scores were aggregated monthly and compared with EDWI---a survey-based business sentiment index---published by the Government of Japan.  The result demonstrated the validity of our proposed approach with a strong positive correlation ($r=0.888$).  The correlation became even higher ($r=0.937$) when compared to the EWDI calculated for survey respondents with industry-related occupations.  This result implies that S-APIR computed from the Nikkei reflects business sentiment in industries more strongly.  It should be also emphasized that S-APIR does not require a costly survey and can be computed much more frequently than monthly EWDI without a time lag.

\item Filtering by our outlier detection model (one-class SVM) worked to remove news texts irrelevant to business sentiment and increased $r$ by 2.1\%.  The effect of filtering was greater for LSTM-BiRNN, which indicates that our fine-tuned BERT is more robust in cases where news texts are noisy, i.e., containing irrelevant texts.  Also, the effect of filtering was greater for general newspapers, increasing $r$ by 10.7\%.

\item Contrary to the intuition that LSTM autoencoder would have an advantage over one-class SVM due to its memory mechanism, the latter was found to be more effective than the former for this relatively simple task to filter out irrelevant texts.  In fact, LSTM autoencoder slightly deteriorated the performance of nowcasting business sentiment index.

\end{itemize}
  
\subsection*{RO2: To analyze the properties of the S-APIR index in comparison with various representative economic indicators and discuss its implications.}
\label{sec:ro2}

\begin{itemize}
\item A comparison with macroeconomic indicators, GDP and DFM, showed that S-APIR captures the macroeconomy as a whole and is slightly ahead of them when there are major shocks. The result implies that S-APIR contains information about the economic outlook and sentiment of economic agents, such as consumers and sellers.

\item To investigate the implication above, S-APIR was compared with economic indicators representing economic outlook and sentiment; specifically, CCI for the consumer side and PMIs for the sales side toward consumers.  It was found that they showed a strong correlation and that S-APIR was particularly consistent with CCI and PMIs on major shocks. The result confirms that S-APIR well captures consumer activity forecasts.

\item To further investigate whether S-APIR precedes \textit{actual} consumption activities, it was compared with SCI from both demand-and supply-side statistics and CAI from only supply-side statistics. During most of the large economic events (e.g., the financial crisis and the COVID-19 pandemic), S-APIR was found to be ahead of the indicators. Thus, S-APIR can be considered as a suitable leading index of actual consumption activity.
\end{itemize}

\subsection*{RO3: To quantify the effects of several notable events on business sentiment and illustrate how it could benefit economists and policymakers.}
\label{sec:ro3}

\begin{itemize}
\item We analyzed the effects of several events on business sentiment by distributing a sentiment score of a sentence to its constituent words and by adding them up for each word (or phrase) representing an event.  It was empirically shown that the analysis could help us examine what factors cause economic fluctuations for a specific period.  One of the examples regarding foreign tourists (``\Ja{インバウンド}'') exemplified the value of the S-APIR index for policymakers in order to assess the negative effect of the COVID-19 pandemic on tourism.  Also, it was shown that the results were robust as to whether sentiment scores were distributed equally or proportionally to attention rollout~\citep{abnar20:_quant_atten_flow_trans}.
\end{itemize}

\section{Conclusions}
\label{sec:conclusions}

This paper reported our work to develop a new business sentiment index, called S-APIR.  The main contribution of this work is threefold: Firstly, we proposed an approach to capturing business sentiment based on news texts and empirically validated it in comparison with an existing survey-based index.  Secondly, we thoroughly studied the properties of the proposed index.  Lastly, we illustrated how the predicted business sentiment can be used by policymakers and economists when it was broken down into individual events.  The following describes, more specifically, the contribution from methodological, theoretical, and practical viewpoints.

The methodological contribution is that we devised an effective framework composed of outlier detection and prediction models.  The former used one-class SVM to identify news texts related to the economy and the latter was a BERT model fine-tuned on Economy Watchers Survey to predict the sentiment score of input news text.   Another contribution is that we proposed  an approach to analyzing the effect of an event represented by an individual/compound word on business sentiment.

Next, the theoretical contribution is that business sentiment was shown to be accurately measured by news articles instead of a traditional, large-scale survey.  Our evaluation using the Nikkei Newspaper demonstrated that S-APIR had a strong positive correlation with an existing business sentiment index, EWDI, up to 0.937.  Also, the result suggested that the S-APIR index more accurately reflects business sentiment in industries.

Then, the practical contribution is that S-APIR does not require a costly survey and can be computed much more frequently than monthly EWDI with almost no time lag.  From the comparison with other business conditions indicators, it was revealed that S-APIR is useful as a leading index of actual consumption activity especially during major economic events such as the global financial crisis and the COVID-19 recession.  Also, it could help us examine what factors cause economic fluctuations for specific periods.   With several example events, such as ``Tokyo Olympics'', it was demonstrated that S-APIR can be useful for economists or policymakers to measure the impact of any event of their interest on business sentiment over time to promptly respond to, if any, their negative effects when necessary.

The findings of this study, however, should be considered in the light of the following limitations:

\begin{itemize}
\item Both the outlier detection and sentiment analysis models are learned from the past Economy Watchers Survey responses and thus potentially suffered from new words unknown to the models.  To keep up with the latest events, these models need to be regularly updated every time new survey data are available.
  
\item Similarly, while the usefulness of the S-APIR index was demonstrated, the evaluation was done retrospectively on the historical data.  Ideally, it should be evaluated by prospective users on ongoing events with a real-time system where the S-APIR index is dynamically updated as breaking news comes in.

\item Predicted values of the S-APIR index depend on news texts we feed to the model. Currently, we use the Nikkei newspaper, which resulted in a strong correlation with EWDI, but feeding a different newspaper yields a different result as witnessed in Section~\ref{sec:general-newspaper}. The difference may come from the different coverage of different newspapers but we do not know exactly if that is the case.  For example, it might be caused by different political stances or tones of different newspapers.  We plan to investigate it further in future work.

\end{itemize}

We are currently working on automatically collecting online news and applying our models to nowcast daily business sentiment and are planning to provide a temporal analysis tool to be used by economists.

\section*{Acknowledgments}
This work was conducted partly as a research project ``Development and application of new business sentiment index based on textual data'' at APIR and was partially supported by MEXT, Japan; and JSPS KAKENHI \#18K11558, \#20H05633, and \#21K13301. We thank Hideo Miyahara, Hiroshi Iwano, Yuzo Honda, Yoshihisa Inada, and Akira Nakayama for their support.


\begin{thebibliography}{55}
\expandafter\ifx\csname natexlab\endcsname\relax\def\natexlab#1{#1}\fi
\providecommand{\url}[1]{\texttt{#1}}
\providecommand{\href}[2]{#2}
\providecommand{\path}[1]{#1}
\providecommand{\DOIprefix}{doi:}
\providecommand{\ArXivprefix}{arXiv:}
\providecommand{\URLprefix}{URL: }
\providecommand{\Pubmedprefix}{pmid:}
\providecommand{\doi}[1]{\href{http://dx.doi.org/#1}{\path{#1}}}
\providecommand{\Pubmed}[1]{\href{pmid:#1}{\path{#1}}}
\providecommand{\bibinfo}[2]{#2}
\ifx\xfnm\relax \def\xfnm[#1]{\unskip,\space#1}\fi
\bibitem[{Abnar \& Zuidema(2020)}]{abnar20:_quant_atten_flow_trans}
\bibinfo{author}{Abnar, S.}, \& \bibinfo{author}{Zuidema, W.}
  (\bibinfo{year}{2020}).
\newblock \bibinfo{title}{Quantifying attention flow in transformers}.
\newblock In {\it \bibinfo{booktitle}{Proceedings of the 58th Annual Meeting of
  the Association for Computational Linguistics}\/} (pp.
  \bibinfo{pages}{4190--4197}).
\bibitem[{{Abu Farha} \& Magdy(2021)}]{farha21:_arabic}
\bibinfo{author}{{Abu Farha}, I.}, \& \bibinfo{author}{Magdy, W.}
  (\bibinfo{year}{2021}).
\newblock \bibinfo{title}{A comparative study of effective approaches for
  {Arabic} sentiment analysis}.
\newblock {\it \bibinfo{journal}{Information Processing \& Management}\/},
  {\it \bibinfo{volume}{58}\/}, \bibinfo{pages}{102438}.
\bibitem[{Aiba \& Yamamoto(2018)}]{aiba18eng}
\bibinfo{author}{Aiba, Y.}, \& \bibinfo{author}{Yamamoto, H.}
  (\bibinfo{year}{2018}).
\newblock \bibinfo{title}{Data science and new financial engineering}.
\newblock {\it \bibinfo{journal}{Business Observation}\/},  {\it
  \bibinfo{volume}{81}\/}, \bibinfo{pages}{30--41}.
\newblock \bibinfo{note}{In Japanese}.
\bibitem[{Arias et~al.(2014)Arias, Arratia \&
  Xuriguera}]{arias14:_forec_twitt_data}
\bibinfo{author}{Arias, M.}, \bibinfo{author}{Arratia, A.}, \&
  \bibinfo{author}{Xuriguera, R.} (\bibinfo{year}{2014}).
\newblock \bibinfo{title}{Forecasting with {Twitter} data}.
\newblock {\it \bibinfo{journal}{ACM Transactions on Intelligent Systems and
  Technology}\/},  {\it \bibinfo{volume}{5}\/}.
  \DOIprefix\doi{10.1145/2542182.2542190}.
\bibitem[{Behera et~al.(2021)Behera, Jena, Rath \& Misra}]{behera21:_co_lstm}
\bibinfo{author}{Behera, R.~K.}, \bibinfo{author}{Jena, M.},
  \bibinfo{author}{Rath, S.~K.}, \& \bibinfo{author}{Misra, S.}
  (\bibinfo{year}{2021}).
\newblock \bibinfo{title}{{Co-LSTM}: Convolutional {LSTM} model for sentiment
  analysis in social big data}.
\newblock {\it \bibinfo{journal}{Information Processing \& Management}\/},
  {\it \bibinfo{volume}{58}\/}, \bibinfo{pages}{102435}.
\bibitem[{Blei et~al.(2003)Blei, Ng \& Jordan}]{Blei:2003:LDA:944919.944937}
\bibinfo{author}{Blei, D.~M.}, \bibinfo{author}{Ng, A.~Y.}, \&
  \bibinfo{author}{Jordan, M.~I.} (\bibinfo{year}{2003}).
\newblock \bibinfo{title}{Latent dirichlet allocation}.
\newblock {\it \bibinfo{journal}{The Journal of Machine Learning Research}\/},
  {\it \bibinfo{volume}{3}\/}, \bibinfo{pages}{993--1022}.
\bibitem[{Bollen et~al.(2011)Bollen, Mao \& Zeng}]{bollen11:_twitt}
\bibinfo{author}{Bollen, J.}, \bibinfo{author}{Mao, H.}, \&
  \bibinfo{author}{Zeng, X.} (\bibinfo{year}{2011}).
\newblock \bibinfo{title}{Twitter mood predicts the stock market}.
\newblock {\it \bibinfo{journal}{Journal of Computer Science}\/},  {\it
  \bibinfo{volume}{2}\/}, \bibinfo{pages}{1--8}.
\bibitem[{Chakraborty et~al.(2016)Chakraborty, Venkataraman, Jagabathula \&
  Subramanian}]{chakraborty16:_predic_socio_econom_indic_using_news_event}
\bibinfo{author}{Chakraborty, S.}, \bibinfo{author}{Venkataraman, A.},
  \bibinfo{author}{Jagabathula, S.}, \& \bibinfo{author}{Subramanian, L.}
  (\bibinfo{year}{2016}).
\newblock \bibinfo{title}{Predicting socio-economic indicators using news
  events}.
\newblock In {\it \bibinfo{booktitle}{Proceedings of the 22nd ACM SIGKDD
  International Conference on Knowledge Discovery and Data Mining}\/} (pp.
  \bibinfo{pages}{1455--1464}).
\bibitem[{Chen et~al.(2019)Chen, Dunn, Hood, Driessen \&
  Batch}]{chen19:_off_races}
\bibinfo{author}{Chen, J.~C.}, \bibinfo{author}{Dunn, A.},
  \bibinfo{author}{Hood, K.}, \bibinfo{author}{Driessen, A.}, \&
  \bibinfo{author}{Batch, A.} (\bibinfo{year}{2019}).
\newblock \bibinfo{title}{Off to the races: A comparison of machine learning
  and alternative data for predicting economic indicators}.
\newblock In {\it \bibinfo{booktitle}{Big Data for Twenty-First Century
  Economic Statistics}\/} NBER Chapters.
\newblock \bibinfo{publisher}{National Bureau of Economic Research, Inc}.
\newblock \URLprefix \url{https://ideas.repec.org/h/nbr/nberch/14268.html}.
\bibitem[{Derakhshan \& Beigy(2019)}]{derakhshan19:_sentim}
\bibinfo{author}{Derakhshan, A.}, \& \bibinfo{author}{Beigy, H.}
  (\bibinfo{year}{2019}).
\newblock \bibinfo{title}{Sentiment analysis on stock social media for stock
  price movement prediction}.
\newblock {\it \bibinfo{journal}{Engineering Applications of Artificial
  Intelligence}\/},  {\it \bibinfo{volume}{85}\/}, \bibinfo{pages}{569--578}.
\bibitem[{Devlin et~al.(2019)Devlin, Chang, Lee \& Toutanova}]{devlin19:_bert}
\bibinfo{author}{Devlin, J.}, \bibinfo{author}{Chang, M.-W.},
  \bibinfo{author}{Lee, K.}, \& \bibinfo{author}{Toutanova, K.}
  (\bibinfo{year}{2019}).
\newblock \bibinfo{title}{{BERT}: Pre-training of deep bidirectional
  transformers for language understanding}.
\newblock In {\it \bibinfo{booktitle}{Proceedings of the 2019 Conference of the
  North {A}merican Chapter of the Association for Computational Linguistics:
  Human Language Technologies}\/} (pp. \bibinfo{pages}{4171--4186}).
\newblock \DOIprefix\doi{10.18653/v1/N19-1423}.
\bibitem[{Fang \& Zhan(2015)}]{fang15:_sentim}
\bibinfo{author}{Fang, X.}, \& \bibinfo{author}{Zhan, J.}
  (\bibinfo{year}{2015}).
\newblock \bibinfo{title}{Sentiment analysis using product review data}.
\newblock {\it \bibinfo{journal}{Journal of Big Data}\/},  {\it
  \bibinfo{volume}{2}\/}.
  \DOIprefix\doi{https://doi.org/10.1186/s40537-015-0015-2}.
\bibitem[{Ge et~al.(2020)Ge, Qiu, Liu, Gu \& Xu}]{ge20:_beyon}
\bibinfo{author}{Ge, Y.}, \bibinfo{author}{Qiu, J.}, \bibinfo{author}{Liu, Z.},
  \bibinfo{author}{Gu, W.}, \& \bibinfo{author}{Xu, L.} (\bibinfo{year}{2020}).
\newblock \bibinfo{title}{Beyond negative and positive: {Exploring} the effects
  of emotions in social media during the stock market crash}.
\newblock {\it \bibinfo{journal}{Information Processing \& Management}\/},
  {\it \bibinfo{volume}{57}\/}, \bibinfo{pages}{102218}.
\bibitem[{Giachanou \& Crestani(2016)}]{giachanou16:_like_it_not}
\bibinfo{author}{Giachanou, A.}, \& \bibinfo{author}{Crestani, F.}
  (\bibinfo{year}{2016}).
\newblock \bibinfo{title}{Like it or not: A survey of twitter sentiment
  analysis methods}.
\newblock {\it \bibinfo{journal}{ACM Computing Survey}\/},  {\it
  \bibinfo{volume}{49}\/}.
\bibitem[{Goshima et~al.(2019)Goshima, Takahashi \& Yamada}]{goshima19eng}
\bibinfo{author}{Goshima, K.}, \bibinfo{author}{Takahashi, D.}, \&
  \bibinfo{author}{Yamada, T.} (\bibinfo{year}{2019}).
\newblock \bibinfo{title}{Construction of business news index by natural
  language processing and its application to volatility prediction}.
\newblock {\it \bibinfo{journal}{Financial Research}\/},  {\it
  \bibinfo{volume}{38}\/}.
\newblock \bibinfo{note}{In Japanese}.
\bibitem[{Hochreiter \&
  Schmidhuber(1997)}]{hochreiter97:_long_short_term_memor}
\bibinfo{author}{Hochreiter, S.}, \& \bibinfo{author}{Schmidhuber, J.}
  (\bibinfo{year}{1997}).
\newblock \bibinfo{title}{Long short-term memory}.
\newblock {\it \bibinfo{journal}{Neural Computation}\/},  {\it
  \bibinfo{volume}{9}\/}, \bibinfo{pages}{1735--1780}.
\bibitem[{Jain(2019)}]{haughwout19:_macro}
\bibinfo{author}{Jain, A.} (\bibinfo{year}{2019}).
\newblock \bibinfo{title}{Macro forecasting using alternative data}.
\newblock In \bibinfo{editor}{A.~Haughwout}, \& \bibinfo{editor}{B.~Mandel}
  (Eds.), {\it \bibinfo{booktitle}{Handbook of US Consumer Economics}\/} (pp.
  \bibinfo{pages}{273--327}).
\newblock \bibinfo{publisher}{Academic Press}.
\bibitem[{Khoo \& Johnkhan(2018)}]{khoo18:_lexic}
\bibinfo{author}{Khoo, C.~S.}, \& \bibinfo{author}{Johnkhan, S.~B.}
  (\bibinfo{year}{2018}).
\newblock \bibinfo{title}{Lexicon-based sentiment analysis: Comparative
  evaluation of six sentiment lexicons}.
\newblock {\it \bibinfo{journal}{Journal of Information Science}\/},  {\it
  \bibinfo{volume}{44}\/}, \bibinfo{pages}{491--511}.
\bibitem[{{Kieu} et~al.(2018){Kieu}, {Yang} \& {Jensen}}]{8411269}
\bibinfo{author}{{Kieu}, T.}, \bibinfo{author}{{Yang}, B.}, \&
  \bibinfo{author}{{Jensen}, C.~S.} (\bibinfo{year}{2018}).
\newblock \bibinfo{title}{Outlier detection for multidimensional time series
  using deep neural networks}.
\newblock In {\it \bibinfo{booktitle}{Proceedings of the 19th IEEE
  International Conference on Mobile Data Management}\/} (pp.
  \bibinfo{pages}{125--134}).
\bibitem[{Kondo et~al.(2019)Kondo, Yogosawa, Naruse \& Mori}]{kondo19eng}
\bibinfo{author}{Kondo, H.}, \bibinfo{author}{Yogosawa, M.},
  \bibinfo{author}{Naruse, M.}, \& \bibinfo{author}{Mori, M.}
  (\bibinfo{year}{2019}).
\newblock \bibinfo{title}{Measuring economic trends based on financial
  institution texts}.
\newblock In {\it \bibinfo{booktitle}{Proceedings of the 33rd JSAI}\/}.
\newblock \DOIprefix\doi{10.11517/pjsai.JSAI2019.0_1P2J1302} \bibinfo{note}{in
  Japanese}.
\bibitem[{Kouadri et~al.(2020)Kouadri, Ouziri, Benbernou, Echihabi, Palpanas \&
  Amor}]{kouadri20:_qualit_sentim_analy_tools}
\bibinfo{author}{Kouadri, W.~M.}, \bibinfo{author}{Ouziri, M.},
  \bibinfo{author}{Benbernou, S.}, \bibinfo{author}{Echihabi, K.},
  \bibinfo{author}{Palpanas, T.}, \& \bibinfo{author}{Amor, I.~B.}
  (\bibinfo{year}{2020}).
\newblock \bibinfo{title}{Quality of sentiment analysis tools: The reasons of
  inconsistency}.
\newblock {\it \bibinfo{journal}{Proceedings of the VLDB Endowment}\/},  {\it
  \bibinfo{volume}{14}\/}, \bibinfo{pages}{668--–681}.
  \DOIprefix\doi{10.14778/3436905.3436924}.
\bibitem[{Levenberg et~al.(2014)Levenberg, Pulman, Moilanen, Simpson \&
  Roberts}]{levenberg14:_predic_econom_indic_web_text}
\bibinfo{author}{Levenberg, A.}, \bibinfo{author}{Pulman, S.},
  \bibinfo{author}{Moilanen, K.}, \bibinfo{author}{Simpson, E.}, \&
  \bibinfo{author}{Roberts, S.} (\bibinfo{year}{2014}).
\newblock \bibinfo{title}{Predicting economic indicators from web text using
  sentiment composition}.
\newblock {\it \bibinfo{journal}{International Journal of Computer and
  Communication Engineering}\/},  {\it \bibinfo{volume}{3}\/},
  \bibinfo{pages}{109--115}.
\bibitem[{Li et~al.(2017)Li, Chan, Ou \& Ruifeng}]{li17:_discov}
\bibinfo{author}{Li, B.}, \bibinfo{author}{Chan, K.~C.}, \bibinfo{author}{Ou,
  C.}, \& \bibinfo{author}{Ruifeng, S.} (\bibinfo{year}{2017}).
\newblock \bibinfo{title}{Discovering public sentiment in social media for
  predicting stock movement of publicly listed companies}.
\newblock {\it \bibinfo{journal}{Information Systems}\/},  {\it
  \bibinfo{volume}{69}\/}, \bibinfo{pages}{81--92}.
\bibitem[{Li et~al.(2020)Li, Wu \& Wang}]{li20:_incor}
\bibinfo{author}{Li, X.}, \bibinfo{author}{Wu, P.}, \& \bibinfo{author}{Wang,
  W.} (\bibinfo{year}{2020}).
\newblock \bibinfo{title}{Incorporating stock prices and news sentiments for
  stock market prediction: {A} case of {Hong Kong}}.
\newblock {\it \bibinfo{journal}{Information Processing \& Management}\/},
  {\it \bibinfo{volume}{57}\/}, \bibinfo{pages}{102212}.
\bibitem[{Li et~al.(2014)Li, Xie, Chen, Wang \& Deng}]{li14:_news}
\bibinfo{author}{Li, X.}, \bibinfo{author}{Xie, H.}, \bibinfo{author}{Chen,
  L.}, \bibinfo{author}{Wang, J.}, \& \bibinfo{author}{Deng, X.}
  (\bibinfo{year}{2014}).
\newblock \bibinfo{title}{News impact on stock price return via sentiment
  analysis}.
\newblock {\it \bibinfo{journal}{Knowledge-Based Systems}\/},  {\it
  \bibinfo{volume}{69}\/}, \bibinfo{pages}{14--23}.
\bibitem[{Manevitz \& Yousef(2002)}]{manevitz02:_one_svms_docum_class}
\bibinfo{author}{Manevitz, L.~M.}, \& \bibinfo{author}{Yousef, M.}
  (\bibinfo{year}{2002}).
\newblock \bibinfo{title}{One-class {SVM}s for document classification}.
\newblock {\it \bibinfo{journal}{The Journal of Machine Learning Research}\/},
  {\it \bibinfo{volume}{2}\/}, \bibinfo{pages}{139--154}.
\bibitem[{Manning et~al.(2008)Manning, Raghavan \&
  Sch\"{u}tze}]{Manning:2008:IIR:1394399}
\bibinfo{author}{Manning, C.~D.}, \bibinfo{author}{Raghavan, P.}, \&
  \bibinfo{author}{Sch\"{u}tze, H.} (\bibinfo{year}{2008}).
\newblock {\it \bibinfo{title}{Introduction to information retrieval}\/}.
\newblock \bibinfo{address}{New York, NY, USA}: \bibinfo{publisher}{Cambridge
  University Press}.
\bibitem[{Nguyen et~al.(2017)Nguyen, {Schulte im Walde} \&
  Vu}]{nguyen17:_distin_anton_synon_patter_neural_networ}
\bibinfo{author}{Nguyen, K.~A.}, \bibinfo{author}{{Schulte im Walde}, S.}, \&
  \bibinfo{author}{Vu, N.~T.} (\bibinfo{year}{2017}).
\newblock \bibinfo{title}{Distinguishing antonyms and synonyms in a
  pattern-based neural network}.
\newblock In {\it \bibinfo{booktitle}{Proceedings of the 15th Conference of the
  European Chapter of the Association for Computational Linguistics}\/} (pp.
  \bibinfo{pages}{76--85}).
\bibitem[{Oliveira et~al.(2017)Oliveira, Cortez \& Areal}]{oliveira17}
\bibinfo{author}{Oliveira, N.}, \bibinfo{author}{Cortez, P.}, \&
  \bibinfo{author}{Areal, N.} (\bibinfo{year}{2017}).
\newblock \bibinfo{title}{The impact of microblogging data for stock market
  prediction: Using {Twitter} to predict returns, volatility, trading volume
  and survey sentiment indices}.
\newblock {\it \bibinfo{journal}{Expert Systems with Applications}\/},  {\it
  \bibinfo{volume}{73}\/}, \bibinfo{pages}{125--144}.
\bibitem[{Picasso et~al.(2019)Picasso, Merello, Ma, Oneto \&
  Cambria}]{picasso19:_techn}
\bibinfo{author}{Picasso, A.}, \bibinfo{author}{Merello, S.},
  \bibinfo{author}{Ma, Y.}, \bibinfo{author}{Oneto, L.}, \&
  \bibinfo{author}{Cambria, E.} (\bibinfo{year}{2019}).
\newblock \bibinfo{title}{Technical analysis and sentiment embeddings for
  market trend prediction}.
\newblock {\it \bibinfo{journal}{Expert Systems with Applications}\/},  {\it
  \bibinfo{volume}{135}\/}, \bibinfo{pages}{60--70}.
\bibitem[{Pota et~al.(2021)Pota, Ventura, Fujita \&
  Esposito}]{pota21:_multil_bert}
\bibinfo{author}{Pota, M.}, \bibinfo{author}{Ventura, M.},
  \bibinfo{author}{Fujita, H.}, \& \bibinfo{author}{Esposito, M.}
  (\bibinfo{year}{2021}).
\newblock \bibinfo{title}{Multilingual evaluation of pre-processing for
  {BERT}-based sentiment analysis of tweets}.
\newblock {\it \bibinfo{journal}{Expert Systems with Applications}\/},  {\it
  \bibinfo{volume}{181}\/}, \bibinfo{pages}{115119}.
\bibitem[{Qian et~al.(2020)Qian, Li \& Yuan}]{qian20}
\bibinfo{author}{Qian, Y.}, \bibinfo{author}{Li, Z.}, \& \bibinfo{author}{Yuan,
  H.} (\bibinfo{year}{2020}).
\newblock \bibinfo{title}{On exploring the impact of users’ bullish-bearish
  tendencies in online community on the stock market}.
\newblock {\it \bibinfo{journal}{Information Processing \& Management}\/},
  {\it \bibinfo{volume}{57}\/}, \bibinfo{pages}{102209}.
\bibitem[{Rehman et~al.(2019)Rehman, Malik, Raza \&
  Ali}]{rehman19:_hybrid_cnn_lstm_model_improv}
\bibinfo{author}{Rehman, A.~U.}, \bibinfo{author}{Malik, A.~K.},
  \bibinfo{author}{Raza, B.}, \& \bibinfo{author}{Ali, W.}
  (\bibinfo{year}{2019}).
\newblock \bibinfo{title}{A hybrid {CNN-LSTM} model for improving accuracy of
  movie reviews sentiment analysis}.
\newblock {\it \bibinfo{journal}{Multimedia Tools and Applications}\/},  {\it
  \bibinfo{volume}{78}\/}, \bibinfo{pages}{26597--26613}.
\bibitem[{Ren et~al.(2021)Ren, Dong, Padmanabhan \& Nickerson}]{ren21:_how}
\bibinfo{author}{Ren, J.}, \bibinfo{author}{Dong, H.},
  \bibinfo{author}{Padmanabhan, B.}, \& \bibinfo{author}{Nickerson, J.~V.}
  (\bibinfo{year}{2021}).
\newblock \bibinfo{title}{How does social media sentiment impact mass media
  sentiment? {A} study of news in the financial markets}.
\newblock {\it \bibinfo{journal}{Journal of the Association for Information
  Science and Technology}\/},  {\it \bibinfo{volume}{first published
  online}\/}. \DOIprefix\doi{https://doi.org/10.1002/asi.24477}.
\bibitem[{Seki \& Ikuta(2020)}]{seki20:_s_apir}
\bibinfo{author}{Seki, K.}, \& \bibinfo{author}{Ikuta, Y.}
  (\bibinfo{year}{2020}).
\newblock \bibinfo{title}{{S-APIR:} news-based business sentiment index}.
\newblock In {\it \bibinfo{booktitle}{Proceedings of the 24th European
  Conference on Advances in Databases and Information Systems}\/} (pp.
  \bibinfo{pages}{189--198}).
\bibitem[{Seki \& Ikuta(2021)}]{seki21:_nowcas_busin_sentim_econom_news_artic}
\bibinfo{author}{Seki, K.}, \& \bibinfo{author}{Ikuta, Y.}
  (\bibinfo{year}{2021}).
\newblock \bibinfo{title}{Nowcasting business sentiment from economic news
  articles}.
\newblock {\it \bibinfo{journal}{IPSJ Journal}\/},  {\it
  \bibinfo{volume}{62}\/}.
\newblock \bibinfo{note}{In Japanese}.
\bibitem[{Serrano \& Smith(2019)}]{serrano-smith-2019-attention}
\bibinfo{author}{Serrano, S.}, \& \bibinfo{author}{Smith, N.~A.}
  (\bibinfo{year}{2019}).
\newblock \bibinfo{title}{Is attention interpretable?}
\newblock In {\it \bibinfo{booktitle}{Proceedings of the 57th Annual Meeting of
  the Association for Computational Linguistics}\/} (pp.
  \bibinfo{pages}{2931--2951}).
\bibitem[{Shapiro et~al.(2020)Shapiro, Sudhof \& Wilson}]{shapiro20:_measur}
\bibinfo{author}{Shapiro, A.~H.}, \bibinfo{author}{Sudhof, M.}, \&
  \bibinfo{author}{Wilson, D.~J.} (\bibinfo{year}{2020}).
\newblock \bibinfo{title}{Measuring news sentiment}.
\newblock {\it \bibinfo{journal}{Journal of Econometrics}\/},  {\it
  \bibinfo{volume}{published first online}\/}.
  \DOIprefix\doi{https://doi.org/10.1016/j.jeconom.2020.07.053}.
\bibitem[{Smetanin \& Komarov(2021)}]{smetanin21:_deep_russian}
\bibinfo{author}{Smetanin, S.}, \& \bibinfo{author}{Komarov, M.}
  (\bibinfo{year}{2021}).
\newblock \bibinfo{title}{Deep transfer learning baselines for sentiment
  analysis in {Russian}}.
\newblock {\it \bibinfo{journal}{Information Processing \& Management}\/},
  {\it \bibinfo{volume}{58}\/}, \bibinfo{pages}{102484}.
\bibitem[{Song et~al.(2019)Song, Park \& shik Shin}]{song19:_atten_korean}
\bibinfo{author}{Song, M.}, \bibinfo{author}{Park, H.}, \&
  \bibinfo{author}{shik Shin, K.} (\bibinfo{year}{2019}).
\newblock \bibinfo{title}{Attention-based long short-term memory network using
  sentiment lexicon embedding for aspect-level sentiment analysis in {Korean}}.
\newblock {\it \bibinfo{journal}{Information Processing \& Management}\/},
  {\it \bibinfo{volume}{56}\/}, \bibinfo{pages}{637--653}.
\bibitem[{Stock \& Watson(1989)}]{stock1989}
\bibinfo{author}{Stock, J.~H.}, \& \bibinfo{author}{Watson, M.~W.}
  (\bibinfo{year}{1989}).
\newblock \bibinfo{title}{New indexes of coincident and leading economic
  indicators}.
\newblock {\it \bibinfo{journal}{NBER Macroeconomics Annual}\/},  {\it
  \bibinfo{volume}{4}\/}, \bibinfo{pages}{351--394}.
\bibitem[{Stock \& Watson(1991)}]{stock1991}
\bibinfo{author}{Stock, J.~H.}, \& \bibinfo{author}{Watson, M.~W.}
  (\bibinfo{year}{1991}).
\newblock \bibinfo{title}{A probability model of the coincident economic
  indicators}.
\newblock In \bibinfo{editor}{G.~Moore}, \& \bibinfo{editor}{K.~Lahiri} (Eds.),
  {\it \bibinfo{booktitle}{The Leading Economic Indicators: New Approaches and
  Forecasting Records}\/} (pp. \bibinfo{pages}{63--90}).
\newblock \bibinfo{publisher}{Cambridge University Press}.
\bibitem[{Tay et~al.(2017)Tay, Tuan \&
  Hui}]{tay17:_dyadic_memor_networ_aspec_based_sentim_analy}
\bibinfo{author}{Tay, Y.}, \bibinfo{author}{Tuan, L.~A.}, \&
  \bibinfo{author}{Hui, S.~C.} (\bibinfo{year}{2017}).
\newblock \bibinfo{title}{Dyadic memory networks for aspect-based sentiment
  analysis}.
\newblock In {\it \bibinfo{booktitle}{Proceedings of the 2017 ACM on Conference
  on Information and Knowledge Management}\/} (pp. \bibinfo{pages}{107--116}).
\bibitem[{Tu et~al.(2018)Tu, Yang, Cheung \& Mamoulis}]{tu18:_inves}
\bibinfo{author}{Tu, W.}, \bibinfo{author}{Yang, M.}, \bibinfo{author}{Cheung,
  D.~W.}, \& \bibinfo{author}{Mamoulis, N.} (\bibinfo{year}{2018}).
\newblock \bibinfo{title}{Investment recommendation by discovering high-quality
  opinions in investor based social networks}.
\newblock {\it \bibinfo{journal}{Information Systems}\/},  {\it
  \bibinfo{volume}{78}\/}, \bibinfo{pages}{189--198}.
\bibitem[{Vaswani et~al.(2017)Vaswani, Shazeer, Parmar, Uszkoreit, Jones,
  Gomez, Kaiser \& Polosukhin}]{vaswani17:_atten_all_you_need}
\bibinfo{author}{Vaswani, A.}, \bibinfo{author}{Shazeer, N.},
  \bibinfo{author}{Parmar, N.}, \bibinfo{author}{Uszkoreit, J.},
  \bibinfo{author}{Jones, L.}, \bibinfo{author}{Gomez, A.~N.},
  \bibinfo{author}{Kaiser, L.}, \& \bibinfo{author}{Polosukhin, I.}
  (\bibinfo{year}{2017}).
\newblock \bibinfo{title}{Attention is all you need}.
\newblock In {\it \bibinfo{booktitle}{Proceedings of the 31st International
  Conference on Neural Information Processing Systems}\/} (pp.
  \bibinfo{pages}{6000--6010}).
\bibitem[{Vilares et~al.(2017)Vilares, Gómez-Rodríguez \&
  Alonso}]{vilares17:_univer}
\bibinfo{author}{Vilares, D.}, \bibinfo{author}{Gómez-Rodríguez, C.}, \&
  \bibinfo{author}{Alonso, M.~A.} (\bibinfo{year}{2017}).
\newblock \bibinfo{title}{Universal, unsupervised (rule-based), uncovered
  sentiment analysis}.
\newblock {\it \bibinfo{journal}{Knowledge-Based Systems}\/},  {\it
  \bibinfo{volume}{118}\/}, \bibinfo{pages}{45--55}.
\bibitem[{Watanabe \&
  Watanabe(2014)}]{watanabe14:_estim_daily_inflat_using_scann_data}
\bibinfo{author}{Watanabe, K.}, \& \bibinfo{author}{Watanabe, T.}
  (\bibinfo{year}{2014}).
\newblock \bibinfo{title}{Estimating daily inflation using scanner data: A
  progress report}.
\newblock In {\it \bibinfo{booktitle}{CARF Working Paper Series}\/}
  \bibinfo{number}{CARF-F-342}.
\bibitem[{Xu et~al.(2019)Xu, Meng, Qiu, Yu \&
  Wu}]{xu19:_sentim_analy_commen_texts_based_bilst}
\bibinfo{author}{Xu, G.}, \bibinfo{author}{Meng, Y.}, \bibinfo{author}{Qiu,
  X.}, \bibinfo{author}{Yu, Z.}, \& \bibinfo{author}{Wu, X.}
  (\bibinfo{year}{2019}).
\newblock \bibinfo{title}{Sentiment analysis of comment texts based on bilstm}.
\newblock {\it \bibinfo{journal}{IEEE Access}\/},  {\it \bibinfo{volume}{7}\/},
  \bibinfo{pages}{51522--51532}.
\bibitem[{Yadollahi et~al.(2017)Yadollahi, Shahraki \&
  Zaiane}]{yadollahi17:_curren_state_text_sentim_analy}
\bibinfo{author}{Yadollahi, A.}, \bibinfo{author}{Shahraki, A.~G.}, \&
  \bibinfo{author}{Zaiane, O.~R.} (\bibinfo{year}{2017}).
\newblock \bibinfo{title}{Current state of text sentiment analysis from opinion
  to emotion mining}.
\newblock {\it \bibinfo{journal}{ACM Computing Surveys}\/},  {\it
  \bibinfo{volume}{50}\/}. \DOIprefix\doi{10.1145/3057270}.
\bibitem[{Yamamoto \& Matsuo(2016)}]{yamamoto16eng}
\bibinfo{author}{Yamamoto, Y.}, \& \bibinfo{author}{Matsuo, Y.}
  (\bibinfo{year}{2016}).
\newblock \bibinfo{title}{Sentiment summarization of financial reports by {LSTM
  RNN} model with the {Japan Economic Watcher Survey Data}}.
\newblock In {\it \bibinfo{booktitle}{Proceedings of the 30th JSAI}\/}.
\newblock \bibinfo{note}{In Japanese}.
\bibitem[{Yono \& Izumi(2017)}]{yono17eng}
\bibinfo{author}{Yono, K.}, \& \bibinfo{author}{Izumi, K.}
  (\bibinfo{year}{2017}).
\newblock \bibinfo{title}{Real time sentiment analysis of bank of japan using
  text of financial report and macroeconomic index}.
\newblock In {\it \bibinfo{booktitle}{Proceedings of the 31st JSAI}\/}.
\newblock \bibinfo{note}{In Japanese}.
\bibitem[{Yoshihara et~al.(2014)Yoshihara, Fujikawa, Seki \&
  Uehara}]{yoshihara14:_predic_stock_market_trend_recur}
\bibinfo{author}{Yoshihara, A.}, \bibinfo{author}{Fujikawa, K.},
  \bibinfo{author}{Seki, K.}, \& \bibinfo{author}{Uehara, K.}
  (\bibinfo{year}{2014}).
\newblock \bibinfo{title}{Predicting stock market trends by recurrent deep
  neural networks}.
\newblock In {\it \bibinfo{booktitle}{Proceedings of the Pacific Rim
  International Conference on Artificial Intelligence 2014}\/} (pp.
  \bibinfo{pages}{759--769}).
\bibitem[{Yoshihara et~al.(2016)Yoshihara, Seki \& Uehara}]{yoshihara16:_lever}
\bibinfo{author}{Yoshihara, A.}, \bibinfo{author}{Seki, K.}, \&
  \bibinfo{author}{Uehara, K.} (\bibinfo{year}{2016}).
\newblock \bibinfo{title}{Leveraging temporal properties of news events for
  stock market prediction}.
\newblock {\it \bibinfo{journal}{Artificial Intelligence Research}\/},  {\it
  \bibinfo{volume}{5}\/}, \bibinfo{pages}{103--110}.
\bibitem[{Zhang et~al.(2018)Zhang, Zhang, Wang, Yao, Fang \&
  Yu}]{zhang18:_improv}
\bibinfo{author}{Zhang, X.}, \bibinfo{author}{Zhang, Y.},
  \bibinfo{author}{Wang, S.}, \bibinfo{author}{Yao, Y.}, \bibinfo{author}{Fang,
  B.}, \& \bibinfo{author}{Yu, P.~S.} (\bibinfo{year}{2018}).
\newblock \bibinfo{title}{Improving stock market prediction via heterogeneous
  information fusion}.
\newblock {\it \bibinfo{journal}{Knowledge-Based Systems}\/},  {\it
  \bibinfo{volume}{143}\/}, \bibinfo{pages}{236--247}.
\bibitem[{Zimbra et~al.(2018)Zimbra, Abbasi, Zeng \&
  Chen}]{zimbra18:_state_art_twitt_sentim_analy}
\bibinfo{author}{Zimbra, D.}, \bibinfo{author}{Abbasi, A.},
  \bibinfo{author}{Zeng, D.}, \& \bibinfo{author}{Chen, H.}
  (\bibinfo{year}{2018}).
\newblock \bibinfo{title}{The state-of-the-art in {Twitter} sentiment analysis:
  A review and benchmark evaluation}.
\newblock {\it \bibinfo{journal}{ACM Transactions on Management Information
  Systems}\/},  {\it \bibinfo{volume}{9}\/}.

\end{thebibliography}
\end{document}